%% file: main.tex
\def\year{2021}\relax
\documentclass[letterpaper]{article} % DO NOT CHANGE THIS
\usepackage{aaai21}  % DO NOT CHANGE THIS
\usepackage{times}  % DO NOT CHANGE THIS
\usepackage{helvet} % DO NOT CHANGE THIS
\usepackage{courier}  % DO NOT CHANGE THIS
\usepackage[hyphens]{url}  % DO NOT CHANGE THIS
\usepackage{graphicx} % DO NOT CHANGE THIS
\usepackage{natbib}  % DO NOT CHANGE THIS AND DO NOT ADD ANY OPTIONS TO IT
\usepackage{caption} % DO NOT CHANGE THIS AND DO NOT ADD ANY OPTIONS TO IT
\usepackage{multicol, multirow}
\usepackage[switch]{lineno}
\usepackage{tikz}
\usepackage{amsfonts}
\usepackage{amsmath}
\usepackage{amssymb}
\usepackage{algorithm}
\usepackage{algorithmicx}
\usepackage{algpseudocode}
\usepackage{subfigure}

\input{math_commands}

\newcommand{\trojD}{\Tilde{\mathcal{D}}}
\newcommand{\trigger}{\tau}

\newcommand{\todoc}[2]{{\textcolor{#1}{\textbf{#2}}}}
\newcommand{\todored}[1]{{\todoc{red}{\textbf{[#1]}}}}

\newcommand{\xz}[1]{\todored{XZ: #1}}

\newtheorem{theorem}{Theorem}

\title{\Large \bf  Deep Feature Space Trojan Attack of Neural Networks by Controlled Detoxification}

\author{
    Siyuan Cheng, \textsuperscript{\rm 1}
    Yingqi Liu, \textsuperscript{\rm 1}
    Shiqing Ma, \textsuperscript{\rm 2}
    Xiangyu Zhang. \textsuperscript{\rm 1} \\
}
\affiliations{
    \textsuperscript{\rm 1} Purdue University \\
    \textsuperscript{\rm 2} Rutgers University \\
    cheng535@purdue.edu, liu1751@purdue.edu, sm2283@cs.rutgers.edu, xyzhang@cs.purdue.edu \\
}

\begin{document}
\maketitle
%%%%%%%%%%%%%%%%%%%%%%%%%%%%%%%%%%%%%%%%%%%%%%%%%%%%%%%%%%%%%%%%%%%%%%%%%%%%%%%%

%-------------------------------------------------------------------------------
\begin{abstract}
Trojan (backdoor) attack is a form of adversarial attack on deep neural networks where the attacker provides victims with a model trained/retrained on malicious data. The backdoor can be activated when a normal input is stamped with a certain pattern called trigger, causing misclassification.
Many existing trojan attacks have their triggers being input space patches/objects (e.g., a polygon with solid color) or simple input transformations such as Instagram filters.
These simple triggers are susceptible to recent backdoor detection algorithms. We propose a novel deep feature space trojan attack with five characteristics: {\em effectiveness}, {\em stealthiness},  {\em controllability}, {\em robustness} and  {\em reliance on deep features}. We conduct extensive experiments on 9 image classifiers on various datasets including ImageNet to demonstrate these properties and show that our attack can evade state-of-the-art defense.
\end{abstract}

%-------------------------------------------------------------------------------
\section{Introduction}
\input{intro}

%-------------------------------------------------------------------------------
\section{Related Work}
\input{related_work}

%-------------------------------------------------------------------------------
\section{Defining Feature Space Trojan Attack}
\input{definition}

%-------------------------------------------------------------------------------
\section{Deep Feature Space Trojaning (DFST)}
\input{design}

%-------------------------------------------------------------------------------
\section{Evaluation}
\input{eval}

%-------------------------------------------------------------------------------
\section{Conclusion}
We introduce a new backdoor attack to deep learning models.
Different from many existing attacks, the attack is in the feature space. It leverages a process called controlled detoxification to ensure that the injected backdoor is dependent on deep features instead of shallow ones. Our experiments show that the attack is effective, relatively more stealthy than many existing attacks, robust, and resilient to existing scanning techniques. 

\section{Acknowledgments}
This research was supported, in part by NSF 1901242 and 1910300, ONR N000141712045, N000141410468 and N000141712947, and IARPA TrojAI W911NF-19-S-0012. Any opinions, findings, and conclusions in this paper are those of the authors only and do not necessarily reflect the views of our sponsors.

%-------------------------------------------------------------------------------

\input{main.bbl}
%-------------------------------------------------------------------------------
% Appendix
\input{appendix}

%%%%%%%%%%%%%%%%%%%%%%%%%%%%%%%%%%%%%%%%%%%%%%%%%%%%%%%%%%%%%%%%%%%%%%%%%%%%%%%%
\end{document}

%% file: math_commands.tex
\usepackage{amsmath,amsfonts,bm}

\def\eqref#1{equation~\ref{#1}}
\def\1{\bm{1}}

\def\vm{{\bm{m}}}

\def\vx{{\bm{x}}}

\DeclareMathAlphabet{\mathsfit}{\encodingdefault}{\sfdefault}{m}{sl}
\SetMathAlphabet{\mathsfit}{bold}{\encodingdefault}{\sfdefault}{bx}{n}

\def\sA{{\mathbb{A}}}

\def\sS{{\mathbb{S}}}
\def\sT{{\mathbb{T}}}

\newcommand{\E}{\mathbb{E}}

\newcommand{\R}{\mathbb{R}}

\DeclareMathOperator*{\argmax}{arg\,max}
\DeclareMathOperator*{\argmin}{arg\,min}

%% file: intro.tex
% What is backdoor attack. Existing backdoor attacks. Inject features/objects. The secret is pixel patterns or objects. Simple data-poisoning. As such they tend to overfit the model with simple features.
% Existing defense.
% Our secret is a generator, like a key.

Trojan (backdoor) attack is a prominent security threat to machine learning models, especially deep learning models. It injects secret features called {\em trigger} into a model such that any input possessing such features causes model misclassification. Existing attacks inject such features using additional (malicious) training samples. They vary in the way of generating these samples. For example, data poisoning~\cite{chen2017targeted} assumes the attacker has the access to the training dataset such that he can directly stamp the trigger on some benign samples and set their labels to the target label. Through training, the model picks up the correlation between the trigger and the target label.
Neuron hijacking~\cite{liu2017trojaning} does not assume access to the training set. Instead, it hijacks a small set of neurons in the model and makes them sensitive to the trigger features. It performs model inversion to generate inputs to hijack these neurons.
Reflection attack~\cite{liu2020reflection} uses a filter to inject trigger features by making them look like faded reflection (from glass).
More discussion can be found in the related work section. However, most these attacks do not control the trojan training process. As such, the model tends to overfit on the trigger features and pick up simple features.
In addition, the malicious samples used in many attacks are not stealthy. Manual inspection of the training set can easily disclose the malicious intention. 

Existing defense techniques include detecting malicious inputs (i.e., inputs stamped with triggers) at runtime and scanning models to determine if they have backdoors.
The former cannot decide the malicious identity of a model until malicious inputs are seen at runtime. We hence focus on the latter kind that does not require malicious inputs. {\em Neural Cleanse} (NC)~\cite{wang2019neural} uses optimization to generate a universal input pattern (or trigger) for each output label and observes if there is a trigger that is exceptionally smaller than the others. Note that such a trigger must exist for each label if its size is not bounded (as it could be as large as covering the whole input). ABS~\cite{abs} intentionally enlarges activation values of individual neurons (on benign inputs) to see if the enlarged values can lead to misclassification. If so, such neurons are used to generate a trigger to confirm the malicious identity. {\em Universal Litmus Pattern} (ULP)~\cite{kolouri2020universal} trains on a set of trojaned and benign models to derive a set of universal input patterns. These patterns can lead to different output logits for benign and trojaned models, allowing effective backdoor model classification.
More can be found in the related work section.
These techniques more or less exploit the observation that trojaned models tend to overfit on simple features. Note that the neurons representing these features can be easily enlarged through optimization and hence allow easy trigger reverse engineering and backdoor detection.

We propose a new trojan attack that is stealthier, more difficult to defend, and having configurable attack strength. We call it {\em deep feature space trojan} (DFST) attack.

\smallskip
\noindent
{\bf DFST Attack Model.}
% Like data poisoning, we assume the attacker has access to both the model and the training dataset.
DFST is a poisoning attack, assuming the attacker has access to both the model and the training dataset, and can control the training process. The target label can be any label chosen by the attacker and all the other labels are victims.
The trojaned models will be released to the public just like the numerous benign models. The attacker holds a secret trigger generator. When he wants to launch the attack, he passes a benign input to the trigger generator to stamp an uninterpretable feature trigger, which causes the model to misbehave. The trojaned model behaves normally for inputs that haven't gone through the trigger generator. $\Box$

\smallskip
Instead of having fixed pixel space patches/watermarks or simple color patterns
as the trojan trigger, DFST attack triggers are human uninterpretable features. {\em These features manifest themselves differently at the pixel level for different inputs}. They are injected to the benign inputs through a specially trained generative model called {\em trigger generator} such that humans can hardly tell that the input has been stamped with the trigger features. The secret held by the attacker is the trigger generator instead of fixed pixel patterns/objects.
Although our malicious inputs contain subtle trigger features, a simple trojaning method like data poisoning~\cite{chen2017targeted} that just adds these inputs to the training set may fail to have the model learn the subtle features. Instead, the trojaned model may only extract simple features from the malicious inputs, allowing easy defense. We hence propose a {\em controlled detoxification} technique that restrains the model from picking up simple features. In particular, after the initial data poisoning and the trojaned model achieving a high
attack success rate, we compare the activation values of the inner neurons using the benign inputs and their malicious versions. The neurons that have substantial activation value differences are considered {\em compromised}.
% \xz{Question, why can we not simply invert inputs using the lower level neurons and use them to detoxify. Isn't that easier than the current design?}
Then we use another generative model called {\em detoxicant generator} that has configurable complexity and is able to reverse engineer inputs that can lead to large activation values only
for compromised neurons. These inputs hence contain the features denoted by the compromised neurons and called {\em detoxicant}. These features are usually simple as they can be directly reverse engineered from (compromised) neurons.
% The complexity of the detexicant inputs is determined by the capacity of the generator, which could range from linear filters to multiple-layer fully connected networks with non-linear activation functions. We start with simple generators.
The generated detoxicant inputs are used to retrain the trojaned model so that it can be detoxified from the simple trigger features. The procedure of {\em poisoning and then detoxifying} repeats and eventually the trojaned model can preclude simple trigger features and learns subtle and complex features (as the trigger). The proposed attack has the unique capabilities of controlling attack strength. Specifically, by controlling the complexity of the trigger generator, we can control the abstract level of the feature triggers (e.g., ranging from simple pixel patterns to uninterpretable features); by controlling the complexity of the detoxicant generator, we can force the trojaned model to learn features at different abstract levels that render different detection difficulty. And in the mean time, more complex generators and more abstract trigger features entail longer training time and more rounds of detoxification.

Our contributions are summarized as follows.

\begin{itemize}
    \item We propose deep feature space trojan (DFST) attack. Compared to existing attacks, DFST has the following characteristics: (1) {\em effectiveness} indicated by high attack success rate; (2) {\em stealthiness}, meaning that the accuracy degradation on benign inputs is negligible and it is hard for humans to tell if an input has been stamped; (3) {\em controllability} such that more resource consumption during trojaning leads to more-difficult-to-detect trojaned models; (4) {\em robustness}, meaning that the trigger features cannot be easily evaded by adversarial training of trojaned models; and (5) {\em reliance on deep features}, meaning that the model does not depend on simple trigger features to induce misclassification and is hence difficult to detect.
    
    \item We  formally define feature space trojan attack. Existing pixel space attacks and the proposed DFST are all instances of feature space trojan attack.
    
    \item We devise methods to train the trigger generator and perform controlled detoxification.
    
    \item We develop a prototype to prove the concept. Our evaluation shows that models trojaned by our system have the properties stated earlier. Existing state-of-the-art scanners NC, ABS, and ULP cannot detect the trojaned models. It is available on github~\cite{DFST}. %\xz{make it a reference and cite it} %\xz{put it on github}
\end{itemize}

%% file: related_work.tex
%\siyuan{I will add some more, but I'm afraid that we may exceed the page limit. (limit is 7, but we have 9 now....)}\xz{you need to shorten and summarize, you can even group multiple papers to one kind and use the same sentences. The budget for related work is 3/4 column. For the works that have been discussed in intro, you don't need to discuss any more. Like I said, you shall start the section by "Besides the attacks and defense techniques discussed in introduction, DFST is also related to the following."} \siyuan{Is this OK?}

%In this section, w
We briefly discuss a number of existing trojan attack and defense techniques besides those discussed in introduction.

\noindent
{\bf Trojan (Backdoor) Attacks.}
A number of existing attacks~\cite{chen2017targeted, saha2019hidden, tang2020embarrassingly} are similar to
%Data poisoning attack was first introducted in
data poisoning~\cite{gu2017badnets}, using patch-like triggers. 
%In \cite{gu2017badnets, chen2017targeted, saha2019hidden, tang2020embarrassingly}, 
%the triggers in
%for data poisoning attack 
%are patches like a yellow sticker or a pair of glasses. 
In~\cite{liao2018backdoor}, researchers proposed to trojan neural networks with fixed perturbation patterns which spread all over the input. 
%but the scale for each pixel is limited.
Clean-label attack~\cite{shafahi2018poison, zhu2019transferable,turner2018clean}, different from poisoning attack, plants backdoor without altering the sample labels.
%, by optimization, transfer learning and adversarial sample generation  presented optimization-based methods for crafting poisons, and controlled the classifier behavior using transfer learning. \cite{turner2018clean} conducted a clean-label attack using adversarial examples and GAN-generated data.
Besides, \cite{rezaei2019target} proposed a target-agnostic attack (with no access to target-specific information) based on transfer learning.
\cite{rakin2020tbt} injects backdoor triggers through bit-flipping, while \cite{guo2020trojannet} does that by permuting the model parameters.
\cite{zou2018potrojan} inserts additional malicious neurons and synapses to the victim models.
% \siyuan{@Shiqing. Please give some words about "Dynamic Backdoor Attacks Against Machine Learning Models" and cite this. Thanks.}
\cite{salem2020dynamic} tries to make detection harder by using various dynamic triggers (e.g., different locations, textures) instead of a single static one.
In contrast, our attack is in the feature space, uses a generator to stamp the trigger, and leverages controlled detoxification.

\noindent
{\bf Detection and Defense.} 
STRIP~\cite{strip} detects malicious inputs by adding strong perturbation, which changes the classification result of benign inputs but not malicious inputs.
TABOR~\cite{guo2019tabor} designs a new objective function to find backdoor.
DeepInspect~\cite{chen2019deepinspect} learns the probability distribution of potential triggers from the queried model and retrieves the footprint of backdoors.
\cite{chen2018detecting} leverages activation clustering.
\cite{xu2019detecting} uses meta neural analysis.
\cite{tran2018spectral} uses spectral signatures to identify and remove corrupted inputs.
Fine-pruning~\cite{liu2018fine} removes redundant neurons to eliminate possible backdoors.
\cite{steinhardt2017certified} mitigates attack by constructing approximate upper bounds on the loss across a broad family of attacks.
\cite{doan2019februus} devises an extraction method to remove triggers from inputs and an in-painting method to restore inputs.
%\siyuan{Add three more as required by one reviewer.}
\cite{li2020rethinking} finds malicious inputs by checking accuracy degradation caused by transformations.
\cite{liu2017neural} adopts a similar idea but trains an auto-encoder to obscure injected triggers.
\cite{qiao2019defending} defends backdoors via generative distribution modeling.

%% file: definition.tex
In this section, we formally define feature space trojan attack. Considering a typical classification problem, where the samples $\vx \in \R^d$ and the corresponding label $y \in \{0, 1, \dots, n \}$ jointly obey a distribution $\mathcal{D}(\vx,y)$. Given a classifier $M: \R^d \rightarrow \{0, 1, \dots, n \}$ with parameter $\theta$. The goal of training is to find the best parameter $\argmax_\theta P_{(\vx,y) \sim \mathcal{D}}[M(\vx;\theta)=y]$. Empirically, we associate a continuous loss function $\mathcal{L}_{M,\theta}(\vx,y)$, e.g. cross-entropy, to measure the difference between the prediction and the true label. 
And the goal is rewritten as $\argmin_\theta \E_{(\vx,y) \sim \mathcal{D}}[\mathcal{L}_{M,\theta}(\vx, y)]$.
We use $\mathcal{L}_{M}$ in short for $\mathcal{L}_{M,\theta}$ in the following discussion.
%\ql{ $\1(x)$ is a function, it equals 1 when the condition x is true, it's 0 vice verse. Or it can also be replaced by a loss function. I found it's confusing and replace the function with text} 
%Under the setting of adversarial learning 
\newcommand{\triggerGen}{\sT}
\newcommand{\trojM}{{\mathcal{\overline M}}}
\newcommand{\trojTheta}{{\overline\theta}}
\newcommand{\targetLabel}{y_t}
\newtheorem{definition}{Definition}

\begin{definition}
Trojan attack aims to derive a classifier $\trojM: \R^d \rightarrow \{0, 1, \dots, n \}$ with parameter $\trojTheta$ such that $\argmax_\trojTheta P_{(\vx,y) \sim \mathcal{D}}[\trojM(\vx;\trojTheta)=y\ \text{and}\ \trojM(\triggerGen(\vx);\trojTheta)=\targetLabel ]$, in which $\triggerGen: \R^d \rightarrow \R^d$ is an input transformation that injects a trigger to a natural input sample  $(\vx,y)$ and $y_t$ is the target label.
% A valid trojan attack often has the following additional properties as well.
A trojan attack is {\bf stealthy} if $(\triggerGen(\vx),y)\sim \mathcal{D}$, meaning that the stamped input $\triggerGen(\vx)$ naturally looks like a sample of the $y$ class. In other words, a perfect classifier $\mathcal{M}$ for the distribution $\mathcal{D}$ would have $\mathcal{M}(\triggerGen(\vx);\theta)=\mathcal{M}(\vx;\theta)$.
A trojan attack is {\bf robust} if given perturbation $\boldsymbol\delta \in \sS \subset \R^d$ of a stamped sample $(\triggerGen(\vx), y)$, 
$\trojM(\triggerGen(\vx)+\boldsymbol\delta;\trojTheta)=\targetLabel$. Normally $\sS$ is defined  as an $\ell_p$-ball centered on $0$. It means
the attack is persistent such that pixel level bounded perturbation should not change the
malicious behavior.
\end{definition}

Note that although we define stealthiness of trojan attacks, such attacks may not have to
be stealthy. For example, many existing attacks~\cite{liu2017trojaning, gu2017badnets, chen2017targeted} have pixel patches as triggers that do not look natural (for humans who can be considered a close-to-perfect classifier).
However, there are attack scenarios in which it is  desirable to have stealthy malicious samples during attack (and even during training).
%However, unstealthy attacks can be detected by human eyes or high quality classifiers.
%The stealthiness aspect of trojan 
%attack also dictates that such attack is only feasible when the classifier is not perfect. In practice, perfect classifiers do not exist for real world complex applications.

Trojaned model scanning is defined as follows.

\begin{definition}
Given a pre-trained
model $\trojM$ with parameters $\trojTheta$, and a set of natural
samples $(\vx, y)\sim \mathcal{D}$, determines if there
exists an input transformation function $\triggerGen$ that satisfies the aforementioned properties of trojan attack. 
The presence of the function indicates the model has been compromised. 
\end{definition}

The difficulty of launching a 
successful attack and the strength of the attack vary with the complexity of $\triggerGen$. 
Many existing  attacks use a  patch as the trigger. This 
corresponds to having a simple $\triggerGen$ that replaces part of an input with the patch. As shown in~\cite{abs}, such simple attacks lead to abnormal neuron behaviors and hence easy detection. 
%In the following, we use a simple network to prove
%this must be the case when certain assumptions are satisfied.

\begin{definition}
A feature space trojan attack is a trojan attack in which $\Delta(\triggerGen(\vx), \vx)$ is not a constant. %Given a set of benign training samples $(x_i, y_i)$ with $i=1,\ ...,\ n$, we say $\triggerGen(\vx)$ is well-formed if $\triggerGen(\vx)-\x_i$
\end{definition}

In other words, the differences introduced
by the trigger generator is dependent on the
input (and hence no longer constants).
%in feature space attack. 
Note that although it appears 
that $\triggerGen$ can be any transformation,
 a poorly designed one (e.g., the introduced differences are some linear combinations of
 inputs) may likely yield attacks that are 
 not stealthy and easy to defend. Therefore in the following sections, we introduce how we use
 a generative model as $\triggerGen$.

%% file: design.tex
In this section, we discuss DFST, an instantiation of feature space trojan attack. 

\begin{figure}
\centering
\includegraphics[width=0.35\textwidth]{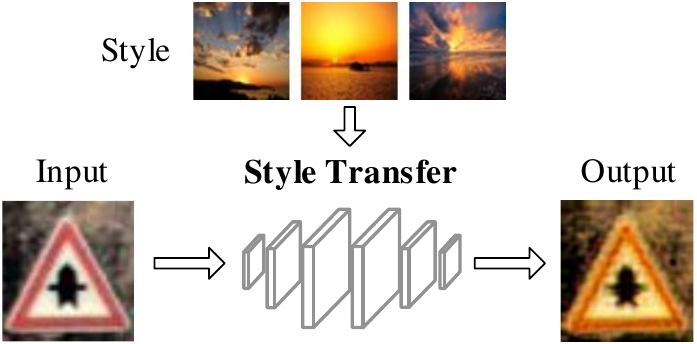}
\caption{Style transfer by CycleGAN
%.\xz{put the two words "Style transfer" to one line to save space and move the three samples in Figure 6 here and remove figure 6}
} \label{fig:CycleGAN_overview}
\end{figure}

\iffalse
\begin{figure*}
\centering
\includegraphics[width=0.7\textwidth]{fig/featureattack.pdf}
\caption{Deep Feature Space Trojaning} \label{fig:trojaning}
\end{figure*}
\fi

\iffalse
\begin{figure*}
\begin{minipage}[t]{0.6\linewidth}
\includegraphics[width=0.7\textwidth]{fig/featureattack.pdf}
\caption{Deep Feature Space Trojaning} \label{fig:trojaning}
\end{minipage}
\hfill
\begin{minipage}[t]{0.3\linewidth}
\includegraphics[width=0.4\textwidth]{fig/poisonous_samples_2.pdf}
\caption{Triggers in detoxification rounds.} \label{fig:poisonous_samples}
\end{minipage}
\end{figure*}
\fi

\begin{figure*}
\centering
\begin{minipage}[c]{.58\textwidth}
  %\center
  \includegraphics[width=0.95\linewidth]{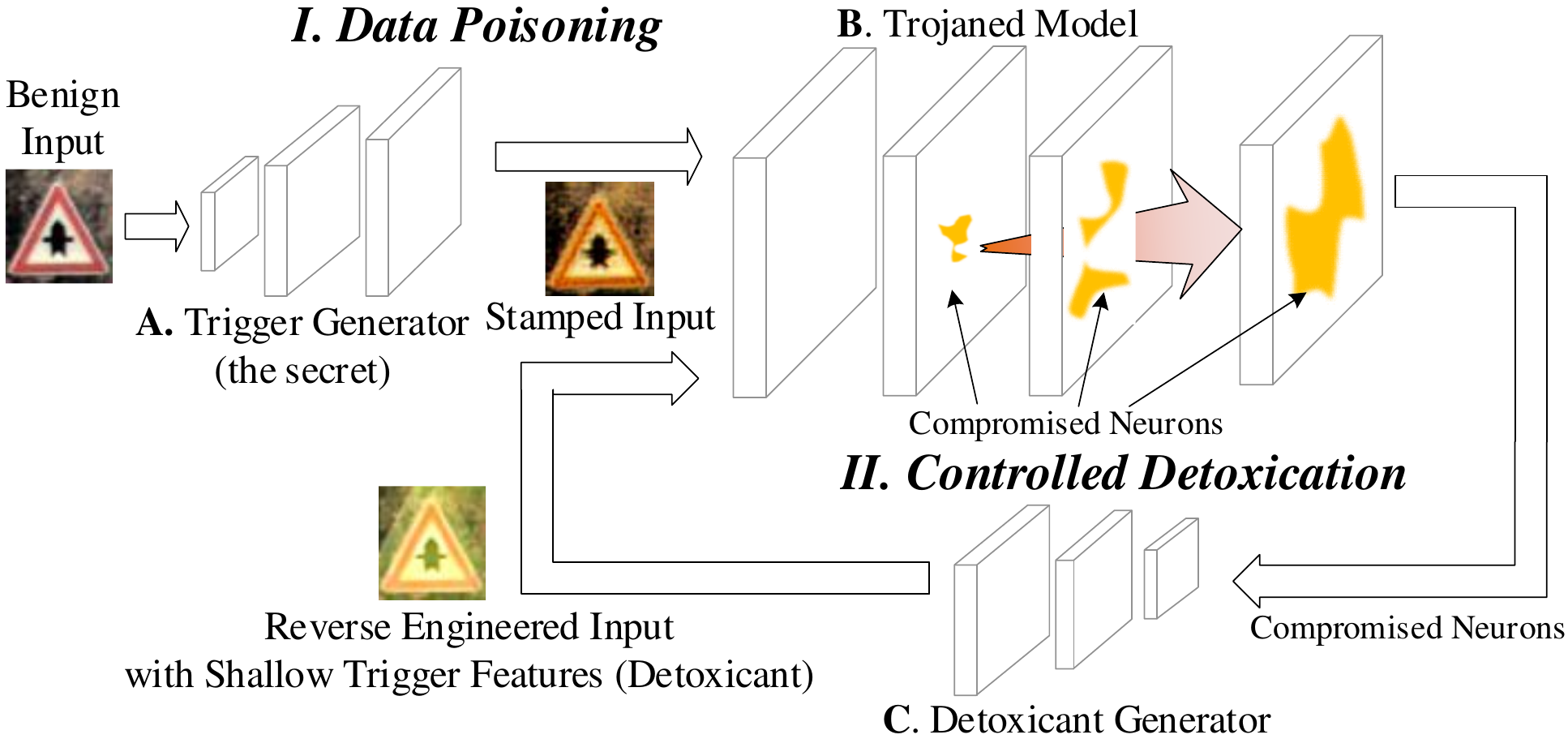}
  \captionof{figure}{Deep Feature Space Trojaning}
  \label{fig:trojaning}
  %\vspace{-0.15in}
\end{minipage}%
$\ \ \ $
\begin{minipage}[c]{.34\textwidth}
  \centering
  \includegraphics[width=1.\linewidth]{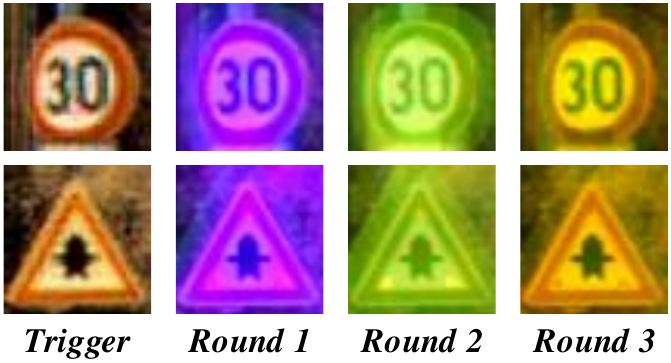}
  \captionof{figure}{Triggers in detoxification rounds}
  
  \label{fig:poisonous_samples}
\end{minipage}
\end{figure*}

\smallskip
\noindent
{\bf Overview.} Our  attack consists of two main steps. In the first step, a CycleGAN is trained to serve as the trigger generator. As shown in Figure~\ref{fig:CycleGAN_overview}, 
%\xz{please put in a diagram for cyclegan training, following a style similar to my diagram. I used visio. The source file is featureattack.vsd} %\siyuan{I have made my version and the source file is CycleGAN.vsd} \xz{your figure is too complex in the overview, I expect to see something similar to the subfigure (A) in your previous version of diagram.jpg. Note that I only used the following two sentence to overview the cyclegan part. I expect the diagram matches the text. The current diagram can be used to explain the details.} \siyuan{Whether the current one is OK?}
%assume the target label is $T$ (i.e., any input
%stamped with the trigger feature ought to be misclassified to $T$ regardless of its original label), 
the generator training procedure takes two sets of images as input: the first is the original training set and the other is a set of images
containing the features that we want to use as trigger, or {\em styles} (such as those commonly appear in sunset shown in the figure), called the {\em style input set}. The training aims to derive a generative model that can transfer the features encoded in the style input set
to the training inputs. Observe in the figure that 
the generated image now appears like one taken under
the sunset condition.
Note that although we use a style transfer CycleGAN model as the 
{\em trigger generator}, as defined in the previous
section, other generators can be used as well. Exploring other generators is left to our future work. 

The second step is to use the trigger generator to trojan the subject
model as shown in Figure~\ref{fig:trojaning}. Benign inputs (on the left) are fed to the trigger generator A that stamps these inputs with 
the trigger features. The stamped inputs, together with the original
benign training inputs, are used in a data poisoning proecedure to
trojan the subject model. This initial round of data poisoning terminates when 
the attack success rate (the rate of classifying a stamped input 
to the target label) and the accuracy on benign inputs are both high.
%According to the discussion in Section~\ref{sec:background}, 
%data poisoning alone often leads to a small number of neurons get compromised 

Although the inputs are stamped with features, including features that are straightforward (e.g., close to pixel patterns) and those that are subtle, 
%according to the discussion in Section~\ref{sec:background}, 
the non-deterministic nature of gradient descent based training dictates that the trojaned model may learn the easy features that lead to high
accuracy.
%\xz{we should have brought up this point earlier in the 
%background section through the limitation discussion and the experiments}. 
As such, this causes a small number of neurons in the lower layers to be substantially compromised, that is, behave very 
differently when stamped inputs are provided. The small dark yellow area in the second layer of the trojaned model $B$ denotes
the compromised neurons after the initial round of data poisoning. To prevent
the model from settling down on simple and shallow features, 
DFST has a unique {\em controlled detoxication} step as part of the
trojaning procedure. Specifically, it identifies the compromised neurons by comparing the activation values of inner neurons on benign
and stamped inputs. A detoxicant generator $C$ takes the identified 
compromised neurons and the original versions of the stamped inputs (which are omitted from the figure for readability), and reverse-engineers inputs that are integration of the provided benign inputs and the (shallow) features denoted by the compromised neurons. 
We call them the {\em detoxicants}. We add these detoxincants to the training set and set their labels to the original correct data labels 
instead of the target label. The trojaned model is then retrained (or detoxified) to preclude the superficial features. 
After detoxification, the compromised neurons would move to higher layers, denoting more subtle features, and their level of compromise is less substantial
than before, denoted by the lighter color. 
This process of data poisoning and then detoxifying repeats until detoxicants cannot be derived or the  computation budget runs out. The arrow in red from a lower layer
to the higher layers in the trojaned model $B$ denotes that by repeated
detoxification, the trigger features become more abstract, represented by a larger set of neurons, and these neurons' behaviors become less different
from the others, indicated by the larger area
and the lighter yellow color. Note this makes detoxicant generation more difficult as well.
%denoting the compromised neurons.

\subsection{Trigger Generator by CycleGAN}
\label{sec:trigger_generation}
Zhu et al.~\cite{zhu2017unpaired} proposed image-to-image translation using {\em Cycle-Consistent Adversarial Networks}, or {\em CycleGAN}.
In this paper, we utilize CycleGAN to train our trigger generator. The trained generator is hence the attacker's secret. Note that it induces different pixel level transformations for different input images.
%For trojan trigger generation, we leverage this model to stimulate real-world style transfer.
%\xz{when you write, try to do top-down writing. First summarize how cycle-gan works. Right now, you  jump right into the details. I do not get how it works at a high level. You organize Figure 4 in the current way for a reason. I do not see the reason. From your text, I only see there are four pieces and there are loss functions. }
Image-to-image translation aims to learn a mapping, $M : A \to B$, between a set of images, $A$, to another set of images, $B$, using training pairs from the two sets.
%\xz{I simplify the description. I do not see the purpose of describing set of sets} 
However, training pairs are usually unavailable. CycleGAN was designed to achieve unpaired image-to-image translation. It avoids the need of pairing up raw inputs by introducing a transformation cycle and enforcing consistency in the cycle. 
The CycleGAN training pipeline consists of two generators ($A\to B$ and $B\to A$) and two discriminators that determine if a sample belongs to $A$ and $B$, respectively. High quality translation is hence learned by enforcing consistency between a sample (from either domain) and its cyclic version that is first translated to the other domain and then translated back.
% Details of CycleGAN are in Appendix A.

%\xz{I find it very strange that there is not section id in this style file. Please make sure you do the right thing}
%\xz{check} \siyuan{I think that's right}

\smallskip
\noindent
{\bf Trigger Generator Construction.}
In our context, the data domain $A$ is the input domain of the subject model while the domain $B$ is the style domain orthogonal to $A$. In this paper, we use a public weather dataset (specifically, sunrise weather) from 
kaggle~\cite{weatheronline}
%kaggle\footnote{\url{https://www.kaggle.com/rahul29g/weatherdataset/notebooks?sortBy=hotness&group=everyone&pageSize=20&datasetId=737827}} 
as $B$. 
%\xz{fix} \xz{I cannot proceed with this section. You need to discuss what are the discriminators. Is $B$ a full dataset of just a few images. Can one choose to have just a few images. Do I need a full dataset? What are the differences of using various setups of $B$. You ignore a lot of interesting details. Imagine I am a reader that tries to reproduce your method. I have a lot of places to fill in. You ought to provide such information}
We use a residual block based auto-encoder for the two generators and a simple CNN with 5 convolutional layers and a sigmoid activation function for the two discriminators. 
%\xz{fill in}
%\siyuan{The discriminator is a simple CNN with binary output. We feed an image to it and it outputs a 0 or 1 to tell whether the input is from the target domain or not. We train the discriminator and the generator at the same time and we optimize both network parameters at every epoch. For example $D_{A}$ judges whether the image is from domain $A$.
%Just like typical GANs, the generators and discriminators are trained. For example, we feed real inputs from $A$ with label $1$, and the generated inputs by $G_{B2A}(B)$, i.e., $\tilde{i}_{A}$, with label $0$ to the discriminator $D_A$ during training. $D_B$ is similarly trained. The generators are updated to fool the discriminators.
%for optimization. For the generator, it tries to fool the discriminator and generates more realistic images like from domain $A$. In this min-max game, both the generator and the discriminator eolve.}
%\siyuan{For the choice of $B$, it's unnecessary to find a large dataset and we just need to collect some representative ones for style transfer. In my experiments, I download 
In our generator training, we used 250 random sunset images from $B$ and  10\% random images from each label in  $A$. 
%The result seems OK I think.... For sure, if you collect a large number of sunrise images in various scenes, the result will be better, but it may be more time-consuming...}
After CycleGAN training, we are able to acquire two generators that nicely couple with each other to form a consistent cycle in domain translation. We use the generator from $A$ to $B$ as the trigger generator. 
%It is also the secret held by the attacker. 
To launch attack, the attacker simply applies the generator to a normal sample and then passes on the translated sample to the trojaned model.
%to induce the malicious behavior.
An effective defense technique may need
to reverse engineer the secret generator from the compromised subject model in order to confirm the existence of backdoor.
%The strength of the secret is hence dependent on the model.

\subsection{Effective Trojaning by Controlled Detoxification}
\noindent
{\bf Limitations of Simple Data Poisoning.}
Many trojan attacks inject their backdoors through {\em data-poisoning}
~\cite{chen2017targeted, gu2017badnets, liu2020reflection, yao2019latent},
%has proposed {\em Data Poisoning} to realize targeted backdoor attack. {\em Data Poisoning}, namely, 
which adds samples stamped with the trigger (e.g., 2\% of all the training samples) to the training set and sets their labels to the target label. We call these samples the {\em malicious samples}. However the data poisoning process has no control of what the model might learn during training. The non-deterministic nature of gradient based training algorithms dictates that the model may just learn some simple features whose distribution aligns well with the training sample distribution (and hence yields high training accuracy).
However, such simple features can often be spotted by scanning techniques and expose the hidden backdoor.

%\xz{I write the following based on the pictures you sent me. Put the pictures in the paper}

\iffalse
\begin{figure}[]
\centering
\includegraphics[width=0.35\textwidth]{fig/poisonous_samples_2.pdf}
\caption{Triggers in detoxification rounds.} \label{fig:poisonous_samples}
\end{figure}
\fi

Consider an example in Figure~\ref{fig:poisonous_samples}, although the malicious samples (in the first column) have the sunset style, the model picks up a simple color setting (demonstrated by the samples in the second column) as the feature that is sufficient to induce the intended mis-classification. 
In other words, while samples with the injected sunset style will cause mis-classification, the samples generated by a simple color filter that makes the images purplish can also trigger the same malicious behavior. The root cause is that the malicious samples have the purplish color scheme as part of its (many) features. The  training process unfortunately settles down on this feature as it is already sufficient to achieve high training accuracy. 
%We want to point out that the purple filter was generated by
The simple feature makes the backdoor easily detectable.
In fact, ABS~\cite{abs} can detect that the model is trojaned as it can reverse engineer a linear transformation trigger equivalent to the purplish filter.
%as a trigger filter, which is a simple linear transformation causing the backdoor behavior.

\begin{figure}
\centering
\includegraphics[width=0.47\textwidth]{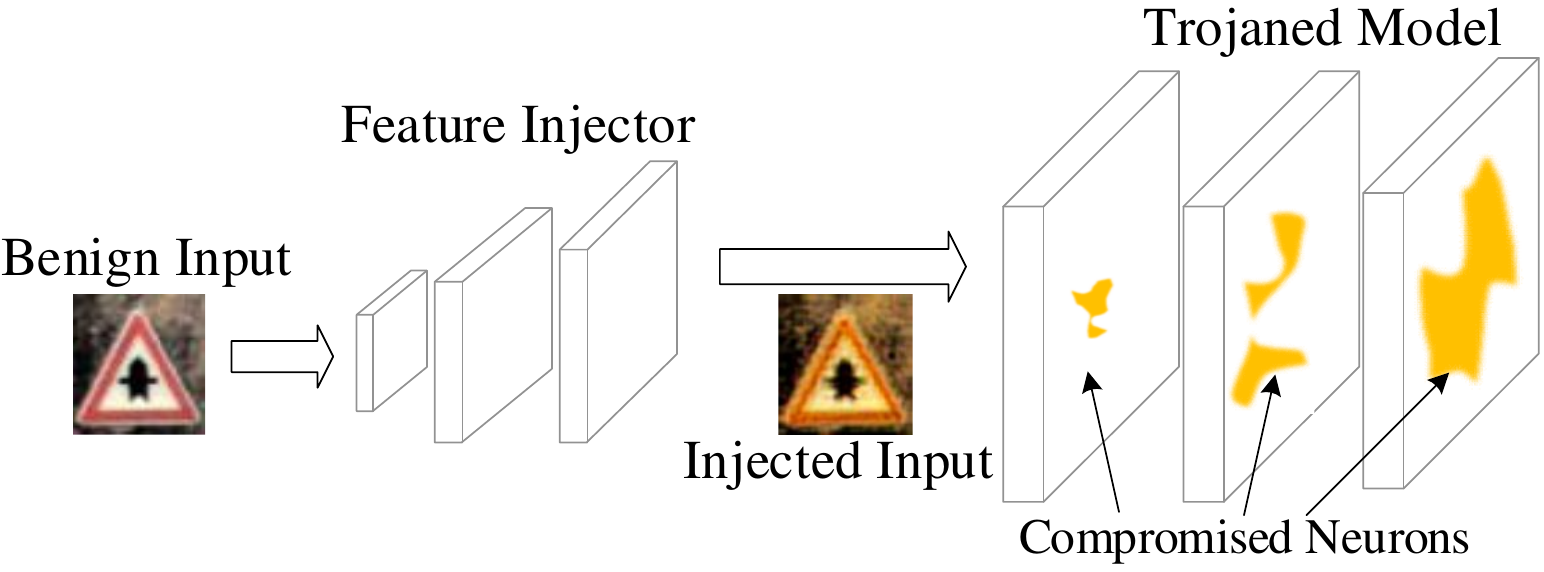}
\caption{Overview of detoxification
}
%.\xz{Lets use a diagram similar to Figure 3, remove the detoxincation part in that figure, replace Trigger generator with feature injector, replace Stampled Input with injected input. remove the red arrow in the trojaned model} \siyuan{Is this OK?}}
\label{fig:detoxification}
%\vspace{-0.15in}
\end{figure}

%\xz{we need a diagram to show detoxification. The diagram shows an input, a simple fully-conneted DNN to reverse engineer, the trojaned CNN model with some layers having highlighted regions to show compromised neurons. Find me on wechat if you are not clear how to draw it. Please read the following text to use the right terms.} \siyuan{How about this one... I think Figure 3 somehow has shown the detoxification process. I have uploaded detoxification.vsd to ./fig folder if you could do some modification.} 
\smallskip
\noindent
{\bf Detoxification Overview.}
%\siyuan{We may somehow shorten this part, which is similar to the overview of DFST as suggested by one reviewer.}
%We hence propose to have iterative detoxification as a needed step to inject backdoors.
%, especially the feature space backdoors. 
%1. identify compromised neurons, explain the whole feature map as a neuron
%2. Model inversion of such neurons using a mask loss
%3. iterative, termination, intuition of termination
%The goal of detoxification 
%is to prevent the model from learning simple features. This is achieved by generating samples containing these simple features and adding them to the training set with their labels set to the original sample labels. These samples are called the {\em detoxicants}.
The detoxicant generation pipeline is shown in Figure~\ref{fig:detoxification}. 
Specifically, we introduce a DNN called {\em feature injector} and use it to model the transformation entailed by the trigger generator.
%\siyuan{I think maybe the following lines are similar to the detoxification part described in the DFST overview, and the feature injector training will be detailed presented in the next subsection, so we may shorten this... and we could save some space.}
%We train the injector as follows. We first identify 
%a set of neurons whose activation values are substantially changed by stamping the trigger. They are called the {\em compromised neurons}. 
%the compromised neurons that denote the trigger features learned by the trojaned model. We then use 
We use the pipeline in Figure~\ref{fig:detoxification} to train the feature injector such that the inputs with injected features can (1) maximize the activation values of the compromised neurons, (2) retain the activation values of other un-compromised neurons (when compared to the original inputs), (3) introduce limited semantic perturbation in the pixel space, and (4) lead to misclassification. 
Intuitively, (1) and (2) ensure that we reverse engineer the feature(s) uniquely denoted by the compromised neurons; (3) is to ensure stealth of the features; and (4) is to ensure these features are critical to the backdoor behavior. 
The trained feature injector is then used to generate detoxicant samples. 
%In this paper, the detoxicants are 2\% of the original training set (while the poisonous samples are 5\%). %\xz{check} \siyuan{I poison the training set with about 5\% images and detoxicants about 2\%}
The detoxification process is iterative as the model may pick up another set of simple features after we preclude one set of them.
%does not imply the model must pick up complex and subtle features. That is, the model may 
The process terminates when the feature injector training cannot converge with a small loss, meaning we cannot find simple features.
In Figure~\ref{fig:poisonous_samples}, the samples in the second column are those generated by the feature injector in the first round of detoxification.
Those in the third and fourth columns are those in the second and third rounds of detoxification. Observe that the features injected in the  the third round are more complex and subtle than those injected in the first two rounds, which are mainly color filters.
%\siyuan{I think these lines are important and we should keep.}
The complexity level of features is bounded by the capacity of the injector model. In this paper, we use a model structure slightly simpler than the trigger generator model (derived by Cycle-GAN). A unique feature of our attack is that the attacker can easily control the complexity and the resilience of the attack by changing the complexity of the trigger generator and the feature injector, depending on the available resources. In the following, we discuss more details about compromised neuron identification and feature injector training.

\subsection{Identifying Compromised Neurons}
\begin{algorithm}
\caption{Compromised Neuron Identification}
\label{Alg:Neuron Selection}
\small
\begin{algorithmic}[1]
\Function{Identify\_Neuron}{$i$, $i_p$,
                          $M, \lambda, \gamma$}
\State $compromised\_neurons=[ ]$
\For{$l$ in $M.layers$}
\State $layer\_v=M(i)[l][:l.neurons]$
\State $max\_v=\max\_value(layer\_v)$
\For{$n$ in $l.neurons$}
\State $troj\_v = M(i_p)[l][n]$
\State $benign\_v = M(i)[l][n]$
\State $\delta = troj\_v - benign\_v$
\If{$\delta > \lambda \cdot max\_v \ \&\& \ \delta > \gamma \cdot benign\_v$}
\State $compromised\_neurons.append(n)$
\EndIf
\EndFor
\EndFor
\State \Return{$compromised\_neurons$}
\EndFunction
\end{algorithmic}
\end{algorithm}
%To train the feature injector, we need to identify compromised neurons.
Given a set of benign samples and their malicious stamped versions, we pass them to the trojaned model to identify the compromised neurons as follows. A neuron is compromised {\em if (1) its activation value for a malicious sample is substantially different from that for the corresponding benign sample and (2) the activation value should be of importance.}
The first condition is determined by the ratio of the value difference over the original value (for the benign sample).
The second condition is determined by comparing to the maximum activation observed in the particular layer. Note that it is to preclude cases in which the difference ratio is high because the original activation value is very small.

Algorithm~\ref{Alg:Neuron Selection} describes the procedure. $M$ denotes the (trojaned) model; $i$ denotes a subset of original samples while $i\_p$ denotes their malicious versions; $\lambda$ and $\gamma$ denote two hyper-parameters. Lines 3-5 compute the maximum activation value $max\_v$ in a layer.
Lines 6-11 first compute the activation value elevation of a neuron $n$, represented by $\delta$, and then determine if $n$ is compromised by the conditions at line 10, that is,
checking if $\delta$ denotes a reasonable fraction of $max\_v$ and hence important and if $\delta$ denotes substantial change over the original value.
The algorithm is for a fully connected layer. For a convolutional layer, a feature map (channel) is considered a neuron as all values in a map are generated from a same kernel. As such, lines 7 and 8 compute the sum of all the values in a feature map.

\subsection{Training Feature Injector}
%\siyuan{I cut this part to save some space since I have put the demo codes on github. And 'vspace' command is not allowed, said in the beginning of 'main.tex'..., so I remove them.}
%\siyuan{
The feature injector is a shallow auto-encoder based on U-net and details of its structure can be found in the github repository.
Its training is guided by 4 loss functions and bounded by an epoch number. 
Algorithm~\ref{Alg:RE} presents the process. $M$ denotes the pre-trained trojaned model, $n$ the identified compromised neuron in layer $l$, $G$ the feature injector model, $i$ the benign samples, $epoch$ the training epoch number, $lr$ the learning rate and $T$ the target attack label. Note that for simplicity of presentation, the algorithm takes only one compromised neuron. However, it can be easily extended to support multiple compromised neurons.
The training loop is in lines 4-15. At line 5, $i'$ denotes the sample with the feature(s) injected. Lines 6-10 denote the four loss functions. The first one (line 6) is the activation value of the compromised neuron (on the feature injected input) and our goal is to maximize it, which explains the negative weight of $f_1$ at line 11. 
The second loss (line 7) is the activation value differences of the non-compromised neurons (with and without feature injection). We want to minimize it and hence its weight is positive at line 11. The third loss (line 9) is the SSIM (or {\em Structural Similarity}) score~\cite{wang2004image} %\xz{fix} \siyuan{OK} 
which measures the perceptional similarity between two images. 
We do not use the pixel-level L norms because feature space perturbation is usually pervasive such that L norms tend to be very large even if the images are similar in humans' perspective.
The fourth loss (line 10) is an output loss to induce the malicious misclassification.

\begin{algorithm}[t]
\caption{Training Feature Injector
%\xz{I changed your algorithm, I move the while statement, I believe you got it wrong in the old version. I also added the cross-entropy loss for mis-classification, could you fill it in? Yingqi found that adding the loss is quite important}
%\siyuan{Is this right?}\xz{I guess I should not say cross-entropy loss, it is more like an output loss to ensure that we get the target label, it is like checking the attack success rate} \siyuan{I add $T$ as the target label. Do you mean to use cross entropy to ensure $G$ generates triggers aiming the target label? Is the current right one OK? $M(i')[T]$ denotes the probability of the generated trojan image $i'$ belonging to the target label $T$ and we try to raise this probability to $1$.}
}
\label{Alg:RE}
\small
\begin{algorithmic}[1]
\Function{Train\_Feature\_Injector}{$M, l, n, G, i$,
                                       $epoch, lr, T$}
\State $initialize(G.weights)$
\State $t = 0$
\While{$t < epoch$}
\State $i' = G(i)$
\State $f_{1} = M(i')[l][n]$
\State $f_{2} = M(i')[l][:n]$
%\State \qquad \  $ + %Model(imgs\_Troj)[l][n+1:]$
 + $M(i')[l][n+1:]$

\State \qquad \  $ - M(i)[l][:n] - M(i)[l][n+1:]$
\State $f_3 = SSIM(i, i')$
\State $f_4 = - \log(M(i')[T])$
\State $cost = - w_{1} \cdot f_{1} + w_{2} \cdot f_{2}- w_{3}\cdot f_3 + w_4\cdot f_4 $
\State $\Delta G.weights = \frac{\partial cost}{\partial G.weights}$
\State $G.weights = G.weights - lr \cdot \Delta G.weights$
\State $i = i + 1$
\EndWhile
\State \Return{$G$}
\EndFunction
\end{algorithmic}
\end{algorithm}

%for successful attack. misclassification). \xz{please fill in the loss} \siyuan{Is this one OK?}

%illustratin
%SSIM scores focus on more st  

%Line 2 first initializes the weights of the generator and line 3 generate the initial trojaned images by $G$. Line 4-8 calculates two costs, $f_{1}$ meaning the activation value of the $n$ and $f_{2}$ meaning the activation value change of other neurons in the same layer $l$. The loop in line 9-16 iteratively optimize the weights of generator to derive an effective trigger. Finally we collect all the generator weights for the next testing part.

%\xz{old writing below and will be removed}
%\siyuan{I think the description of SSIM and how to test the trigger effectiveness may be missed if we remove all of the content below...}
%\xz{trigger effectiveness is replaced with the output loss. SSIM is fine as long as we cite the paper}

%poison the training dataset with a trojaned subset, which consists of trojaned images (at least 1 percent of training images added with a special trigger) and their corresponding target labels, usually only one target label for all trojaned images. Similarly, we leverage {\em Data Poisoning} to install the backdoor into the model, where our trigger is not a simple stamp or accessory but real-world style transfer.

%% file: eval.tex
%In this section we evaluate DFST on 8 classification applications with 4 different model structures, NiN, VGG, ResNet32, ResNet50 and 3 different datasets, CIFAR-10, GTSRB, VGG-Face. 
%Here we define our attack scenario that the attacker has the pre-trained model $M$ and full access to the whole training images $X_{y}$. The attacker then trains a Cycle GAN model to inject the trigger into about 2-percent of the training images through real-world style transfer and entitle them with the target label $X^{\prime}_{y_{t}}$, in our experiment $y_{t}=Label 0$. After using data-poisoning retraining to install the backdoor into $M$, we apply detoxification to finetune it for several times, typically 3 rounds in our experiment and we derive the final model $M^{T}$, which has two functions: 1) $P(M^{T}(X_{y})=y) \to 1$ and 2) $P(M^{T}(X^{\prime}_{y_{t}})=y_{t}) \to 1$ and

We answer the following research questions:  
\begin{itemize}
\item {\bf (RQ1)} Is DFST an effective attack?
\item {\bf (RQ2)} Is DFST stealthy?
\item {\bf (RQ3)} Is detoxification effective?
%in precluding simple backdoor features?
\item {\bf (RQ4)} Can DFST evade existing scanning techniques?
\item{\bf (RQ5)} Is DFST robust?
\end{itemize}
%Our system is available at~\cite{...}. \xz{put the code (even just the compiled code) and some data samples on an anonymous site} %\siyuan{I'm sorry that AAAI said "Submissions should not contain pointers to supplementary material on the web, as this may violate both blind review and the policy that submissions cannot be changed after they are made available to reviewers." Maybe we could submit some codes as supplementary material.}
%Besides the attac success rate, we also evaluate DFST in the following metrics, 
%four aforementioned characteristics: 1) Stealthiness, 2) Controllability, 3) Robustness and 4) Smoothness.

\subsection{Experiment Setup}
Our evaluation is on 9 pre-trained classification systems:
%downloaded from~\cite{...}, \xz{fix} \siyuan{I trained all the systems by myself...} 
NiN, VGG, and ResNet32 on CIFAR-10 and GTSRB, VGG and  ResNet50 on VGG-Face, and ResNet101 on ImageNet. 
%for each function and characteristic mentioned before. In the following, we describe some preparations for the attack process.
%\smallskip
%\noindent
%{\bf Dataset Preprocessing}
% The details of dataset preprocessing  can be found in Appendix B. We will release our system upon publication.
%\xz{move it to Appendix}

%-----------------------------------------------------------------------
\subsection{(RQ1) Is DFST an effective attack?}
\begin{figure}
\centering
\begin{minipage}[b]{0.45\textwidth}
\includegraphics[width=1\textwidth]{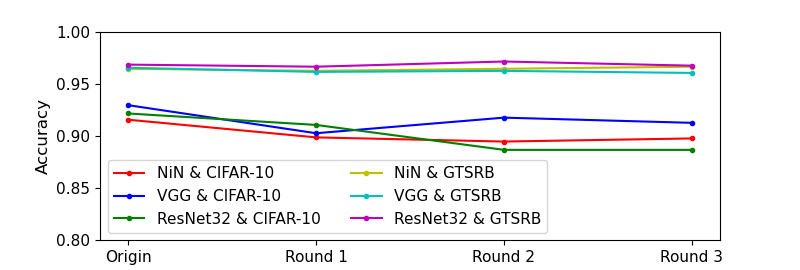} \\
\includegraphics[width=1\textwidth]{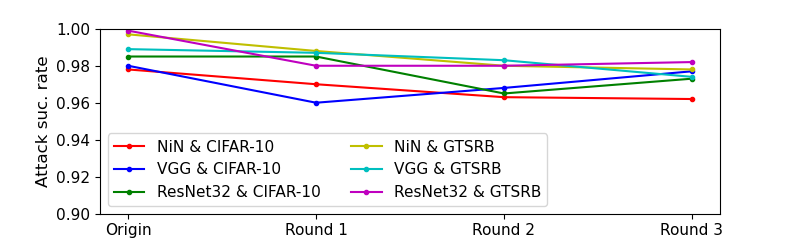}
\end{minipage}
\caption{Accuracy variation (upper one) and attack success rate variation (lower one) during detoxification for NiN, VGG, and ResNet32 on CIFAR-10 and GTSRB.
%\xz{change the lower figure's y axis to attack suc. rate} \siyuan{OK}
%\xz{again, your figures are taking too much space, you can divide the legend to two subparts} \siyuan{Is this OK?}
}
\label{fig:trojan_acc_var}
\end{figure}

%In this section, we test the attack effectiveness of DFST attack. As
We evaluate the effectiveness of DFST  by measuring its accuracy on benign samples and its attack success rate on malicious samples  transformed by the trigger generator. 
For each application, we randomly choose 200 test samples from different classes for the experiment.
%\xz{fix} 
%expected, the trojaned model should remain high testing accuracy on original testing set and trojaned testing set, which is composed of trigger-installed images with target label. 
Table~\ref{tab:accuracy} presents the results after data poisoning. 
Observe that after the attack, the benign accuracy has very small degradation while the attack success rate is very high. %several testing accuracy on different model and dataset before detoxification. The third row shows the testing accuracy on benign testing set of intact pre-trained models while the fourth and fifth rows show the testing accuracy on benign and trojaned testing set of models after data-poisoning. All the high accuracy illustrates the effectiveness beforehand.
Figure~\ref{fig:trojan_acc_var} shows the  variations during detoxification for NiN, VGG, and ResNet32 on CIFAR-10 and GTSRB. %\xz{explain the model and dataset} \xz{I would suggest to split to two graphs, one showing the accuracy variation for all 8 (hence you will have 8 lines), and the other showing the attack success rate for all 8} \siyuan{Is this one OK?} 
Observe that the accuracy and attack success rate  have only small fluctuations and both remain high.
Our experiments are conducted on GeForce RTX 2080 Ti.
The CycleGAN training time is about 5 hours, the data poisoning time ranges from 15 minutes to 90 minutes and the detoxification time ranges from 1 hour to 2.5 hours. Details are elided.
%\xz{fix, some rough numbers are good enough} \siyuan{OK.} 
Note that these are one-time cost. 

%process and the light fluctuation implies the high attack effectiveness through the end.

\begin{table}[]
\centering
\small
\caption{Test accuracy before and after data poisoning} \label{tab:accuracy}
\begin{tabular}{ccccc}
\multirow{2}{*}{Dataset}  & \multirow{2}{*}{Model}
& \multirow{2}{*}{Before} & \multicolumn{2}{c}{After} \\ \cline{4-5}
& & & Benign & Malicious \\ \hline
%& \begin{tabular}[c]{@{}c@{}}Before \end{tabular} & \begin{tabular}[c]{@{}c@{}}Trojaned Model\\ Benign Accuracy\end{tabular} & \begin{tabular}[c]{@{}c@{}}Trojaned Model\\ Attack Accuracy\end{tabular} \\ \hline
         & NiN & 0.914 & 0.916 & 0.978 \\
CIFAR-10 & VGG & 0.925 & 0.930 & 0.980 \\
         & ResNet32 & 0.918 & 0.922 & 0.985 \\
\hline
         & NiN & 0.963 & 0.967 & 0.997 \\
GTSRB    & VGG & 0.973 & 0.966 & 0.989 \\
         & ResNet32 & 0.967 & 0.969 & 0.999 \\
\hline
\multirow{2}*{VGG-Face} & VGG & 0.831 & 0.807 & 0.852 \\
~         & ResNet50 & 0.819 & 0.794 & 0.920 \\
\hline
ImageNet & ResNet101 & 0.912 & 0.904 & 0.990 \\
\hline
\end{tabular}
\end{table}

\begin{table}[]
\centering
\caption{Test accuracy of malicious samples on the original pre-trained models
%\siyuan{I add some other to compare with DFST}
} \label{tab:stealth}
\small
\begin{tabular}{ccccc}
Dataset  & Model     & DFST & Instagram & Reflection \\ \hline
         & NiN       & {\color[HTML]{FE0000} 0.55} & 0.35      & 0.41       \\
CIFAR-10 & VGG       & {\color[HTML]{FE0000} 0.61} & 0.31      & 0.51       \\
         & ResNet32  & {\color[HTML]{FE0000} 0.58} & 0.30      & 0.45       \\ \hline
         & NiN       & {\color[HTML]{FE0000} 0.58} & 0.16      & 0.44       \\
GTSRB    & VGG       & {\color[HTML]{FE0000} 0.88} & 0.35      & 0.47       \\
         & ResNet32  & {\color[HTML]{FE0000} 0.79} & 0.42      & 0.43       \\ \hline
\multirow{2}*{VGG-Face} & VGG       & {\color[HTML]{FE0000} 0.81} & 0.56      & 0.55       \\
~         & ResNet50  & {\color[HTML]{FE0000} 0.74} & 0.64      & 0.57       \\ \hline
ImageNet & ResNet101 & 0.65 & {\color[HTML]{FE0000} 0.68}      & 0.67       \\ \hline
\end{tabular}
\end{table}

%-----------------------------------------------------------------------

\subsection{(RQ2) Is DFST stealthy?}
%In this section, we evaluate the stealthiness of our attack from both human eyes and models' perspective.

\begin{figure}
\centering
\includegraphics[width=0.45\textwidth]{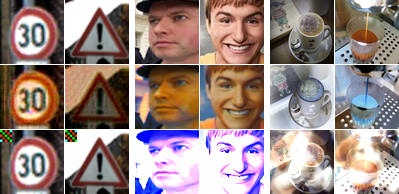}
\caption{Samples on GTSRB, VGG-Face and ImageNet before (the first row) and after injecting the DFST triggers (the second row), and after injecting  triggers by existing attacks, including patch, Instagram filter and reflection (the third row). We use their default settings for existing attacks.}
\label{fig:style_transfer}
\end{figure}

\smallskip
\noindent
%{\bf Visual Effect.}
Figure~\ref{fig:style_transfer} shows a set of samples before and after injecting the DFST triggers, and after injecting watermark/patch~\cite{chen2017targeted,liu2017trojaning}, Instagram filter~\cite{abs}, and refelction~\cite{liu2020reflection}.  We argue that DFST triggers look more natural than those by existing attacks.
%\xz{you may put another row of samples that are from data poisoning, yingqi's work, and reflection. Pick the ones that look unnatural}
%on several image sets, where the first row is the original images while the second row is the style-transferred(trigger-installed) images. We can not tell the trojaned images with the trigger is abnormal, unlike watermark which can be easily detected by naked eyes.
%\smallskip
%\noindent
%{\bf Model output.} 
%Besides visual effect, we also test the stealthiness from the models' perspective, which is the output of the model when inputting trojaned images, style- transferred images and we call the process {\em Stealth Test}. Stealth test computes the accuracy of testing set composed of style-transferred images with their original correct labels.
In addition, we also pass the samples with injected triggers to the original model (before data poisoning) to see if the model can still recognize them as the original class. We use the same test sets in the previous experiment.
Table~\ref{tab:stealth} presents the results. Observe that while the test accuracies degrade, the model can still largely recognize the DFST's transformed images, %indicating the level of stealth of our attack.
%\siyuan{I
indicating DFST has good
stealthiness.
%in the feature space.
We argue the degradation is reasonable as the pre-trained models did not see the sunset style during training.
%test accuracy is about $60\%$ on CIFAR-10 and $80\%$ on GTSRB and VGG-Face, both are reasonable high that  enough to prove that most prime features used for correct classification are kept and that our trigger is stealthy according to the model.

%Possible reason for the somehow low stealth test accuracy on CIFAR-10 is that the pixel value distribution of CIFAR-10 is not very typical(exact), which cause large difficulty for Cycle GAN model to do style transfer between two domains since the discriminator could not fully determine the exact distribution of CIFAR-10 images. However, for GTSRB and VGG-Face, it's distribution is more like a whole, so it's easier to train a Cycle GAN model, thus leading to high stealth test accuracy.

\begin{figure}
\centering
\includegraphics[width=0.4\textwidth]{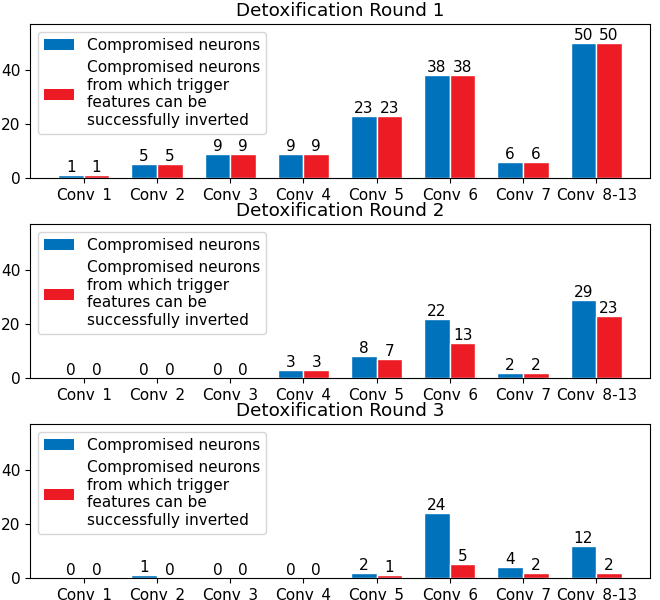}
\caption{Model internals after individual rounds of detoxification for VGG on GTSRB.}
\label{fig:detoxification_V_G}
\end{figure}

%-----------------------------------------------------------------------

\subsection{(RQ3) Is detoxification effective in precluding simple backdoor features?} 
%\subsection{Controllability}
%In this part, we evaluate the controllability of our DFST. We assume the detoxification operation will 1) reduce the number of compromised neurons, and 2) push compromised neurons to higher layers, which means simple trojan features in lower layers will be eliminated and leave behind more abstract features. At the same time the testing accuracy on both the benign and trojan image set will still remain high enough to perform as an effective classification system with functional backdoor installed. So, the detoxification operation grants the attack controllability, which means the more time and resources the attacker consume, the more difficulties he will add to users' detection process.

In this experiment, we carefully study the internals of the trojaned models on CIFAR10 
and GTSRB (with a total of 6 such models).
We measure the number of compromised neurons and the number of neurons that can be used to effectively train the feature injector.
%(i.e., allow small training loss).
To simplify the setup, we train the feature injector using the compromised neurons one by one and measure the attack success rate of using the samples with injected features.
Note that a high attack success rate means that we are able to reverse engineer important features (that can trigger the backdoor behaviors) from the compromised neuron. Figure~\ref{fig:detoxification_V_G} shows the results for VGG on GTSRB. Each sub-figure shows the results for one round of detoxification. 
Inside each sub-figure, two bars are presented for each hidden layer. The blue bar presents the number of compromised neurons for that layer
%\xz{I am confused by the data in the bar charts. What does it mean by a neuron, the whole feature map or a single activation value in the map? I recall that Yingqi used whole feature maps.} \siyuan{Yeah.. Here the neuron means one kernel in the convolutional layer, and the activation value is the sum of the feature map after the convolution of that kernel.}
and the red bar presents the number of compromised neurons that can be used to successfully train the feature injector (i.e., yielding a comparable attack success rate with the real triggers). %\xz{you only show the barcharts for one model, does another one look better? the trend of moving up is more obvious?} \siyuan{I draw all of them today and attach them to the appendix. Please have a look.}\xz{lets just use the current one} \siyuan{Do we need to refer to appendix for the remaining figures?}\xz{no, remove them}
Observe that with the growth of detoxification rounds, the number of compromised neurons is decreasing, especially in the shallow layers. The number of compromised neurons that can be used to derive features is decreasing too, in a faster pace. It indicates 
although there are still compromised neurons, they tend to couple with other neurons to denote more complex/abstract features such that optimizing individual neurons fails to invert the corresponding features. The graphs for other models are similar and hence elided. To summarize, detoxification does suppress the simple features.

%-----------------------------------------------------------------------

\subsection{(RQ4) Can DFST evade scanning techniques}
We evaluate our attack against three state-of-art backdoor scanners, ABS~\cite{abs}, Neural Cleanse (NC)~\cite{wang2019neural}, and ULP~\cite{kolouri2020universal}. Our results show that none of them is effective to detect models attacked by DFST. Details can be found in the repository~\cite{DFST}. 
%\siyuan{I refer to my repository, and we could give a link of the full version on the repository.}
%\xz{the tables and figures and texts are not in a good order in the appendix. For example, a figure/table referenced in a piece text is not close to the text as we moved the text. Please move the figures/tables to their right place} \siyuan{Is this OK?}
%which have been introduced and explained in the background section~\ref{sec:background}.

\subsection{(RQ5) Is DFST robust?}
To study robustness, we conduct three experiments. The first is to study if the injected backdoors can survive two popular adversarial training methods FGSM~\cite{goodfellow2014explaining} and PGD~\cite{madry2017towards}.
%to the 6 trojaned and detoxified models on CIFAR10 and GTSRB. 
%In the adversarial training, we preclude all the malicious or detoxicant samples and only start with the original benign samples. This is to simulate the situation in which normal users harden pre-trained models.
%We use the default settings for the two methods (i.e., $\epsilon = 0.2$ for FGSM and $l_\infty= 5/255$ for PGD).
%\xz{check} \siyuan{It's right.}
In the second experiment, we use {\em randomized smoothing}~\cite{cohen2019certified}
to study the certified (radius) bound  and accuracy of a trojaned model on both the benign and the malicious samples.
In the third one, we perform several spacial and chromatic transformations~\cite{li2020rethinking} to test the degradation of attack success rate (ASR) and check DFST's robustness against pre-processing defending.
The results show that DFST is robust. Details can be found in the repository~\cite{DFST}.

%\siyuan{I refer to my repository, and we could give a link of the full version on the repository.}

%identify the robustness bound for the original pre-train models (on benign inputs) and the trojaned models (on inputs stamped with triggers). 
%\xz{check if I say the right thing} \siyuan{We may need to add (radius) to "bound" since the term in the table is radius..}

%% file: appendix.tex
\clearpage
\section{Appendix}

\subsection{A. Details of CycleGAN} 
\begin{figure}
\centering
\includegraphics[width=0.47\textwidth]{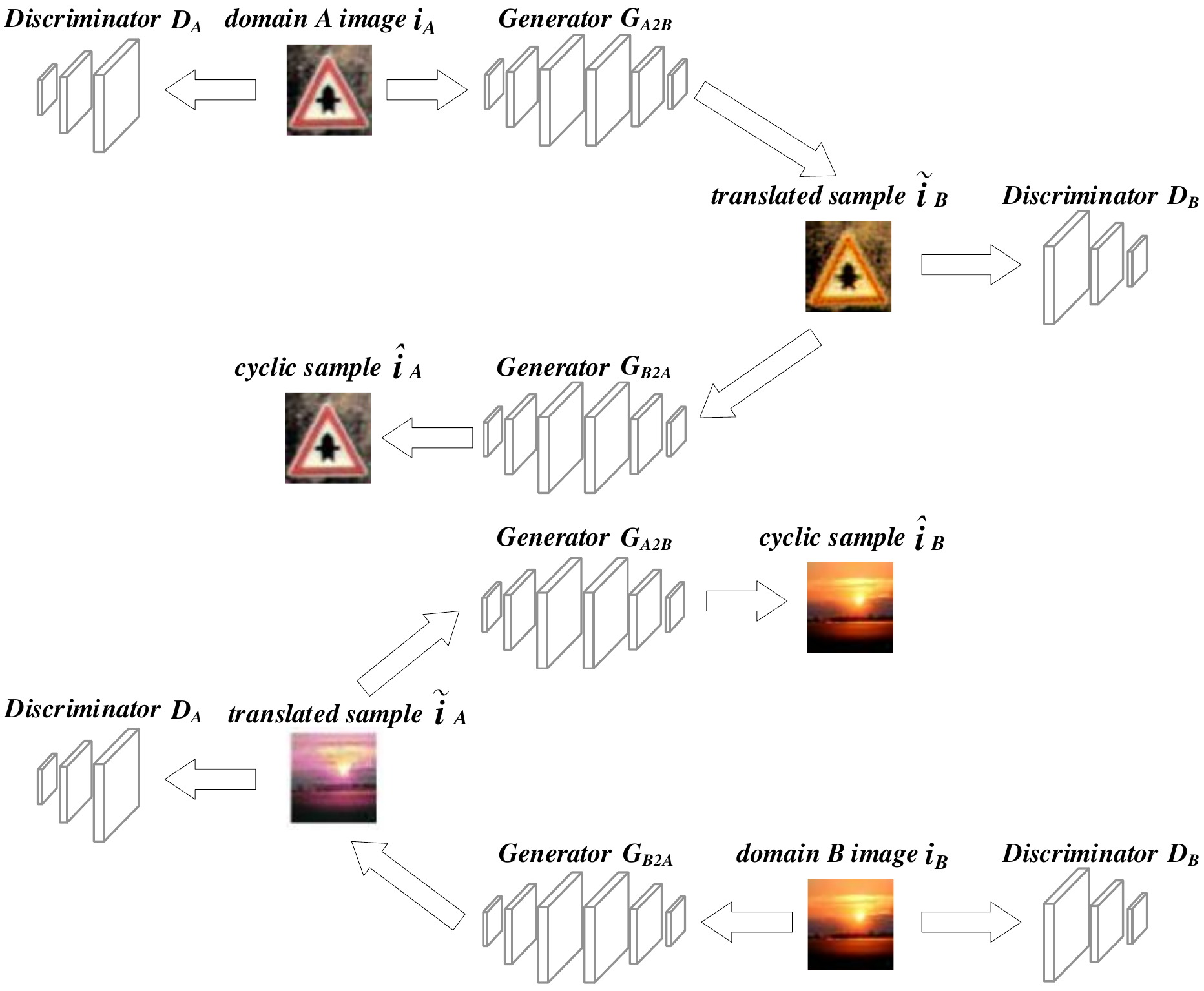}
\caption{CycleGAN structure} \label{fig:CycleGAN structure}
\end{figure}

Given the two (orthogonal) input image domains, $A$ and $B$, CycleGAN learns a mapping $G : A \to B$ such that $G(A) \approx B$ using adversarial loss. Besides, it also learns the inverse mapping $F : B \to A$. In order to generate high-quality mappings, it leverages two training cycles, the first enforces $F(G(A)) \approx A$ and the other $G(F(B)) \approx B$, driven by cycle consistency losses. %and introduces a cycle consistency loss to enforce the inverted images to resemble the original input images,  and . 
Figure~\ref{fig:CycleGAN structure} shows the structure of CycleGAN, with the two cycles next to each other vertically, and proceeding in opposite directions. %\xz{change the figure to make terms consistent}
%Unlike traditional GANs, t
There are two generators, $G_{A2B}$ and  $G_{B2A}$, and two discriminators $D_{A}$ and $D_{B}$ in the structure.
%\xz{the symbols int the figure need to be consistent with those in the text. Please use subscripts as well}
Suppose we have two image domains, $A$ and $B$. $G_{A2B}$ translates an image from domain $A$, denoted as $i_A$, to an image in domain $B$, denoted as a {\em translated sample} $\tilde{i}_B$, while $G_{B2A}$ the opposite.
We denote the input generated by applying %$G_{A2B}$ on $i_A$ as $i'_B$,  
$G_{B2A}$ on $\tilde{i}_{B}$ as $\hat{i}_{A}$, called the {\em cyclic sample}.
%, and vice versa. 
The two discriminators determine if a sample is in the respective domains.
%translates an image from domain $B$, that is $Input_{B}$, to domain $A$, denoted as $Generated_{A}$. Plus we call the reconstructed images $Generator_{A2B}(Generated_{A})$ as $Cyclic_{B}$ and $Generator_{B2A}(Generated_{B})$ as $Cyclic_{A}$. $Discriminator_{A}$ judges whether the input belongs to domain $A$ while $Discriminator_{B}$ judges whether the input belongs to domain $B$. 
%For convenience, in the following, we use $G$ to refer to a generator, $D$ to a discriminator, $a$ to inputs from domain $A$, and $b$ to inputs from domain $B$.

%\xz{that is what I meant by first explaining the intuition}
Three kinds of loss functions are used in training. The first one is the typical adversary loss or GAN loss that ensures the generated samples fall into a specific domain. The second is the {\em cycle consistency loss} that ensure the two generators are appropriately inverse to each other. The last one is the {\em identity loss} which ensures that if an input in the target domain is provided to a generator, the generator has no effect on the input.

%Adversarial losses are applied to both mapping functions to enforce the generated images similar to the images from the target domain\xz{You shall explain the intuition of these loss functions}:
The following presents the GAN losses for the two respective generators. We use $a$ and $b$ to denote samples from $A$ and $B$, respectively. 

{\small
\begin{equation}
\begin{aligned}
    &\mathcal{L}_{GAN}(G_{A2B}, D_{B}, A, B) = \mathbb{E}_{b \sim \mathcal{P}_{data}(b)}[\log D_{B}(b)] \\
    &+ \mathbb{E}_{a \sim \mathcal{P}_{data}(a)}[\log (1-D_{B}(G_{A2B}(a)))],
\end{aligned}
\end{equation}
\begin{equation}
\begin{aligned}
    &\mathcal{L}_{GAN}(G_{B2A}, D_{A}, B, A) = \mathbb{E}_{a \sim \mathcal{P}_{data}(a)}[\log D_{A}(a)] \\
    &+ \mathbb{E}_{b \sim \mathcal{P}_{data}(b)}[\log (1-D_{A}(G_{B2A}(b)))],
\end{aligned}
\end{equation}
}
%where $G$ acts as the mapping function while $D$ aims to distinguish between the generated images and the real images from the target domain. $G$ tries to minimize this objective against an adversary $D$ that tries to maximize it and they play the min-max game to improve the mapping effect.\xz{I don't understand the disscussion here. Could you intuitively explain (5) and (6)? Doest D function produce a value in [0,1]? What is the meaning of the first E of (5)? Why do you care about the loss of log D(b)?} \siyuan{(1) D function produce a value in [0,1] to express how much the generated image G(x) resembles x; (2) I think E expectation is just used to show the function is applied to all the images in the training set; (3) In the original GAN paper, Goodfellow used log likelihood to express the loss, and it seems that log has a larger domain [-inf, 0] instead of just [0, 1], while during exp, we usually just use L2 to train for ease.}

The cycle consistency loss is to reduce the difference between inputs and their projected versions after a cycle (i.e., after a mapping and then the inverse mapping).

{\small 
\begin{equation}
\begin{aligned}
    \mathcal{L}_{cyc}(G_{A2B}, G_{B2A}) &= \mathbb{E}_{a \sim \mathcal{P}_{data}(a)}[\|G_{B2A}(G_{A2B}(a))-a\|_{1}] \\
    &+\mathbb{E}_{b \sim \mathcal{P}_{data}(b)}[\|G_{A2B}(G_{B2A}(b))-b\|_{1}].
\end{aligned}
\end{equation}
}

The identity loss is defined as follows.
%means that a generator model is expected to output an image without translation when provided an example from the target domain:

{\small
\begin{equation}
\begin{aligned}
    \mathcal{L}_{id}(G_{A2B}, G_{B2A}) &= \mathbb{E}_{a \sim \mathcal{P}_{data}(a)}[\|G_{B2A}(a)-a\|_{1}] \\
    &+\mathbb{E}_{b \sim \mathcal{P}_{data}(b)}[\|G_{A2B}(b)-b\|_{1}].
\end{aligned}
\end{equation}
}

The overall objective function is hence defined as an aggregation of the three aforementioned losses.
%of the original Cycle GAN consists of the above three parts according to the paper:

{\small
\begin{equation}
\begin{aligned}
    \mathcal{L}(G_{A2B}, G_{B2A}, D_{A}, D_{B}) &= \mathcal{L}_{GAN}(G_{A2B}, D_{B}, A, B) \\
    &+ \mathcal{L}_{GAN}(G_{B2A}, D_{A}, B, A) \\
    &+ \alpha \mathcal{L}_{cyc}(G_{A2B}, G_{B2A}) \\
    &+ \beta \mathcal{L}_{id}(G_{A2B}, G_{B2A}),
\end{aligned}
\end{equation}
}

Here $\alpha$ and $\beta$ control the relative importance of three objectives, usually $\alpha=10$ and $\beta=1$, and we aim to solve:

{\small
\begin{equation}
    G_{A2B}^{*}, G_{B2A}^{*} = \arg\min_{G_{A2B}, G_{B2A}} \max_{D_{A}, D_{B}} \mathcal{L}(G_{A2B}, G_{B2A}, D_{A}, D_{B}).
\end{equation}
}

Intuitively, it aims to search for the generator parameters that can minimize the maximum adversarial loss.

\subsection
{B. Dataset Preprocessing}
\begin{itemize}
\item CIFAR-10~\cite{cifar} is a well-known standard image classification dataset with the image size of $32\times32$. It has 10 classes, with 5000 images per class for training and 1000 images per class for testing. For the initial data-poisoning, we randomly select 100 images from all 10 classes, 2\% of the original training set, while for the detoxification rounds, we apply inject features to to 50 images per class (in each round), 1\% of the original training set.
\item  GTSRB~\cite{Stallkamp2012} is another widely used image classification dataset of German traffic signs. It contains 43 classes and about $3.7k$ images. While they are not of the same size, we resize the  images to $48\times48$. Similarly, we take 100 images per class for data-poisoning and 50 images per class for detoxification.
\item VGG-Face~\cite{parkhi2015deep} is a common face recognition dataset that contains $2,622$ identities with 1000 photos each. %For convenience,
We conduct our experiments based on a subset of  20 labels with 500 images per label. Similarly, we resize the images to $224\times224$ and take 50 images per class for data-poisoning and 10 images per class for detoxification.
\item   ImageNet~\cite{deng2009imagenet} is a large object recognition dataset,  containing over 15 millions high-resolution ($224\times224$) images in roughly 22,000 categories. We use a subset with 10 classes and 1000 images per class. 
%\xz{how many were used in poisoning and detoxification?} \siyuan{I remember we have discussed the rate of poisoning and detoxification before, and poisoning data is about
We use 50 images per class for data-poisoning and 20 images per class for detoxification.
%5\% of the training set while detoxification is about 2\%.}
%for our DFST evaluations.
\end{itemize}

% Remaining detoxification illustrations
\iffalse
\begin{figure}
\centering
\includegraphics[width=0.45\textwidth]{fig/detoxification_N_C.png}
\caption{Model internals after individual rounds of detoxification for NiN on CIFAR-10.} \label{fig:detoxification_N_C}
\end{figure}

\begin{figure}
\centering
\includegraphics[width=0.45\textwidth]{fig/detoxification_V_C.png}
\caption{Model internals after individual rounds of detoxification for VGG on CIFAR-10.} \label{fig:detoxification_V_C}
\end{figure}

\begin{figure}
\centering
\includegraphics[width=0.45\textwidth]{fig/detoxification_R_C.png}
\caption{Model internals after individual rounds of detoxification for ResNet32 on CIFAR-10.} \label{fig:detoxification_R_C}
\end{figure}

\begin{figure}
\centering
\includegraphics[width=0.45\textwidth]{fig/detoxification_N_G.png}
\caption{Model internals after individual rounds of detoxification for NiN on GTSRB.} \label{fig:detoxification_N_G}
\end{figure}

\begin{figure}
\centering
\includegraphics[width=0.45\textwidth]{fig/detoxification_R_G.png}
\caption{Model internals after individual rounds of detoxification for ResNet on GTSRB.} \label{fig:detoxification_R_G}
\end{figure}
\fi

\subsection{C. (RQ4) Can DFST evade scanning techniques}
We evaluate our attack against three state-of-art backdoor scanners, ABS~\cite{abs}, Neural Cleanse (NC)~\cite{wang2019neural}, and ULP~\cite{kolouri2020universal}. %which have been introduced and explained in the background section~\ref{sec:background}.

%\subsubsection{ABS Test}

\begin{figure}
\centering
\includegraphics[width=0.3\textwidth]{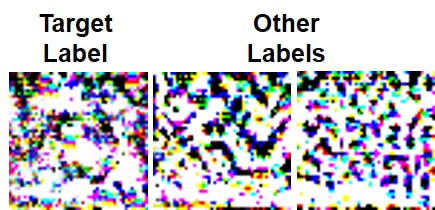}
\caption{Trigger pattern examples by Neural Cleanse. The first is the target label trigger pattern and the other two are for other labels. Note that there is not much size difference.} \label{fig:nc-result}
\end{figure}

\smallskip
\noindent
{\bf Neural Cleanse (NC).}
As mentioned in the introduction, NC uses optimization to find a universal perturbation pattern (called {\em trigger pattern}) for each output label such that applying the pattern to any sample causes the model to classify the label. It considers a model trojaned if the pattern of some label is much smaller than those of the others.
%(and hence a trigger patten).
%Neural Cleanse reverse engineer trojan triggers through end-to-end optimization from the input space to the final label space and it judges whether the model is attacked based on the size of its reverse engineered triggers, the label with extreme small trigger size will be considered target model and the model itself with backdoor. 
It uses a metric called {\em anomaly index} to measure the anomaly level of a trigger pattern. If there is a label with an anomaly index larger than 2, it considers the model trojaned.
%has $>95\%$ probability of being an outlier, affected label.
Table~\ref{tab:nc} shows the NC results. Observe that none of the DFST-attacked models can be detected. The underlying reason is that
the input transformation by our attack is global (i.e., on a whole image) so that the corresponding pixel space trigger pattern is very large.
%detects the model depending on the reverse engineered trigger size, but in our attack, the trigger is a large, whole-image perturbation
Figure~\ref{fig:nc-result} shows some examples of the generated trigger patterns.
%, so neural cleanse loses its effect.

\begin{table}[]
\small
\centering
\caption{Neural Cleanse result} \label{tab:nc}
\begin{tabular}{cccc}
Dataset  & Model  & Anomaly Index & Detected \\ \hline
%& & & (\textgreater{}2 means trojaned) \\ \hline
         & NiN    & 1.629         & $\times$ \\
CIFAR-10 & VGG    & 1.853         & $\times$ \\
         & ResNet32 & 1.790         & $\times$ \\ \hline
         & NiN    & 1.853         & $\times$ \\
GTSRB    & VGG    & 1.529         & $\times$ \\
         & ResNet32 & 1.918         & $\times$ \\ \hline
VGG-Face & VGG    & 1.440         & $\times$ \\
         & ResNet50 & 1.798         & $\times$ \\ \hline
ImageNet & ResNet101 & 1.308 & $\times$ \\ \hline
\end{tabular}
\end{table}

\smallskip
\noindent
{\bf ABS.} ABS improves over NC by using a stimulation analysis to first identify neurons that demonstrate abnormal behaviors when their activation values are enlarged, and then using an optimization technique similar to NC to generate trigger, with the guidance of those neurons. It considers a model trojaned if a small trigger can be generated. It also handles simple filter attacks (e.g., triggering a backdoor by applying a Gaussian Instagram filter to a benign input) by optimizing a kernel
as the trigger. In this case, it considers a model trojaned if applying the kernel (through a sliding window) to the input can cause misclassification.   

\begin{figure}
\centering
\includegraphics[width=0.3\textwidth]{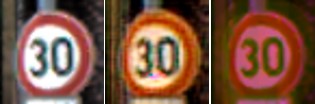}
\caption{Examples to illustrate ABS. From left to right, the original image, the image with our trigger applied, and the image with the trigger by ABS applied. Note that here ABS is applied before detoxification. The trigger by ABS hence induces misclassification.} \label{fig:abs-result}
\end{figure}

%ABS detects backdoor attack through two steps, neuron stimulation analysis to find some compromised neurons candidates and trigger reverse engineer validation to generate the effective trojan triggers and thus prove the model attacked. In this experiment, we search $20$ potential compromised neurons and apply the reverse engineering to detect our attack.
%As mentioned before, detoxification process is the soul of our attack which controls the detection difficulty level. 
Table~\ref{tab:abs} shows the results. 
%The numbers mean XXX \xz{what is 20? what is 1, 4,2 ?} \siyuan{/20 means in each compromised neuron candidate detecting round, we choose 20 neurons with largest activation value elevation. 1, 4, 2 means that among the found 20 compromised neurons, 1, 4, 2 neurons are real compromised neurons and will cause the backdoor, trigger the attack.}
Observe that before detoxification, ABS can detect a few of the trojaned models. But it can detect none after. This is because detoxification suppresses neuron abnormal behaviors. A sample trigger generated by ABS can be found in Figure~\ref{fig:abs-result}. 
%of ABS before and after detoxification process. One can easily see that ABS is effective to find the effective trojan trigger before detoxification since there are several noticeable compromised neurons at this stage and they are easy to be reverse engineered through pooling transformation proposed by ABS. However, since that ABS has no access to the exact trojan pattern and that simple pooling transformation could just reverse engineer less abstract features~\ref{fig:abs-result}, it could just find some most outstanding compromised neurons with shallow perturbation through stimulation and validation, detoxification could easily eliminate them and vitiate ABS's detection.

\begin{table}[]
\small
\centering
\caption{ABS results before and after detoxification. The numbers in parentheses denote the neurons that can be used to reverse engineer triggers.} \label{tab:abs}
\begin{tabular}{cccc}
Dataset  & Model  & Before  & After  \\
& & Detoxification & Detoxification \\ \hline
         & NiN    & $\checkmark$ (1) & $\times$ \\
CIFAR-10 & VGG    & $\checkmark$ (4) & $\times$ \\
         & ResNet32 & $\checkmark$ (2) & $\times$ \\ \hline
         & NiN    & $\checkmark$ (3) & $\times$ \\
GTSRB    & VGG    & $\checkmark$ (5) & $\times$ \\
         & ResNet32 & $\times$ & $\times$ \\ \hline
VGG-Face & VGG    & $\times$ & $\times$ \\
         & ResNet50 & $\times$ & $\times$ \\ \hline
ImageNet & ResNet101 & $\times$ & $\times$ \\ \hline
\end{tabular}
\end{table}

\smallskip
\noindent
{\bf ULP.}
%\xz{put the ulp results here. Don't have to be a lot. Just like the other two.}
ULP trains a number of input patterns from a large number of benign and trojaned models. 
These patterns are supposed to induce different output logits for benign and trojaned models such that they can be used for scanning. ULP is claimed to be model structure agnostic and trigger-type agnostic~\cite{kolouri2020universal}. 
Since it requires a large number of models for training, we use the TrojAI Round 1 training dataset\footnote{\url{https://pages.nist.gov/trojai/docs/data.html}} that consists of 500 benign models and 500 trojaned models (with various model structures and trigger types) to train the patterns. The training accuracy reaches $95\%$. 

%Universal Litmus Patterns (ULPs) are kinds of universal patterns for revealing backdoor attack detection by feeding them to the network and analyzing the output. Simply put, several ULPs and one model classifier are training on a model set (containing a large number of benign and corrupted models of various network structures but with the same input and output shape). For round, ULPs are fed to the model and the output logits values are concatenate as the input for the model classifier, which will give a binary output to tell whether the model is attacked or not. Both the ULPs and the model classifier are trained at the same time. Although the training phase may cost much time, the detection is fast since it requires only a few forward passes through the model. We train our ULPs and model classifier on the TrojAI Round 1 dataset\footnote{\url{https://pages.nist.gov/trojai/docs/data.html}} with a large number of models and test on our
We then apply the trained ULP to the
DFST-attacked VGG and ResNet50 on VGG-Face and ResNet101 on ImageNet. We cannot test on CIFAR-10 or GTSRB as the ULP is trained on high-resolution images ($224*224$)). Table~\ref{tab:ulp} shows the  results. Observe none of the DFST-attacked models can be detected. 
It discloses that the unique DFST attack mechanism makes the ULP {\em trained on  existing models} ineffective. 
Due to the high detoxification cost, acquiring a large set of DFST attacked models for ULP training is presently infeasible, we will leave it to our future work. 

%The reasons are that most of the training trigger patterns are small singular polygons which is far different from our feature space triggers, and that shows a defect for ULP: Detection effectiveness relies much on the similar trigger style between the training and testing samples.\xz{I am worried they may ask that we should train on DFST attacked models. What shall we say?} 
%\siyuan{I think 1) It's hard to collect a large number of benign $\&$ trojaned model for training ULPs, so the ULP method itself may not be that practical and convenient; 2) In the ULP paper, they do not require that the training triggers should be similar to the testing triggers to some extent and they claim ULPs are invariant to various network structures and potentially various triggers. Therefore, we could assume that trained ULPs should have some detecting abilities targeting various triggers, including DFST. 3) Here is a simple evaluation and we could leave the training ULPs on DFST triggers as further work.}

\begin{table}[]
\small
\centering
\caption{ULP result} \label{tab:ulp}
\begin{tabular}{ccc}
Dataset   & Model     & Detected \\ \hline
VGG-Face  & VGG       & $\times$             \\
          & ResNet50  & $\times$             \\ \hline
ImageNet & ResNet101 & $\times$             \\ \hline
\end{tabular}
\end{table}

\subsection{D. (RQ5) Is DFST robust?}
In this section, we study the robustness of DSFT. We conduct three experiments. The first is to study if the injected backdoor can survive adversarial training.
We apply two popular adversarial training methods {\em Fast Gradient Sign Method} (FGSM)~\cite{goodfellow2014explaining} and {\em Projected Gradient Descent} (PGD)~\cite{madry2017towards} to the 6 trojaned and detoxified models on CIFAR10 and GTSRB. 
In the adversarial training, we preclude all the malicious or detoxicant samples and only start with the original benign samples. This is to simulate the situation in which normal users harden pre-trained models.
We use the default settings for the two methods (i.e., $\epsilon = 0.2$ for FGSM and $l_\infty= 5/255$ for PGD).
%\xz{check} \siyuan{It's right.}
In the second experiment, we use {\em randomized smoothing}~\cite{cohen2019certified}
to study the certified (radius) bound  and accuracy of a trojaned model on both the benign and the malicious samples. %identify the robustness bound for the original pre-train models (on benign inputs) and the trojaned models (on inputs stamped with triggers). 
%\xz{check if I say the right thing} \siyuan{We may need to add (radius) to "bound" since the term in the table is radius..}
In the third one, we perform several spacial and chromatic transformations~\cite{li2020rethinking} to test the degradation of attack success rate(ASR) and check DFST's robustness against pre-processing defense.

\begin{figure}
\centering
\includegraphics[width=0.25\textwidth]{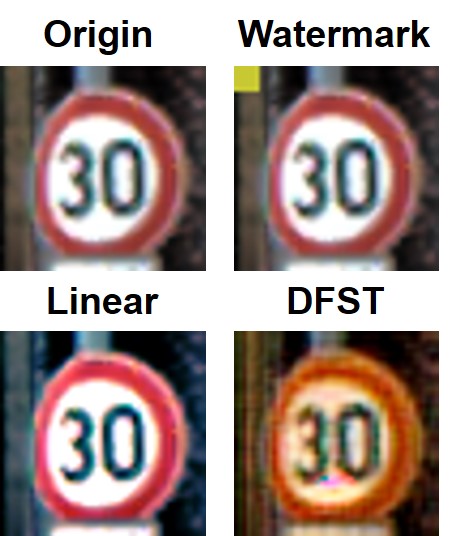}
\caption{Examples of three types of backdoor attack triggers} \label{fig:3-attack}
\end{figure}

For comparison, we perform the two adversarial training methods on three different kinds of backdoor attacks: the pixel-space patch (or watermark) attack as in the original data poisoning paper~\cite{gu2017badnets,chen2017targeted}, the linear filter attack mentioned in ABS~\cite{abs} and our DFST attack.
%We apply both of the adversarial training to trojan attack model of three trigger types, watermark, linear filter and our DFST attack. Watermark trigger is a small monochromic square placed at the corner of the image, linear filter do same linear transformation on the input images, like adjusting pixel value range and changing mean and standard deviation, and our trigger is generated using Cycle GAN to apply real-world style transfer.
Figure~\ref{fig:3-attack} shows trigger examples of the three types of attacks.

Table~\ref{tab:robustness} shows the results of adversarial training. 
We observe that the both the watermark and DFST attacks are resilient to adversarial training, meaning the attack success rate after adversarial training does not degrade, whereas the linear filter attack has substantial (up to 58\%) degradation.
This seems to indicate in the linear filter attack, the trojaned models tend to learn unrobust trigger features~\cite{tsipras2018robustness, ilyas2019adversarial}. %\siyuan{"Towards deep learning models resistant to adversarial attacks" This one?}.\xz{not sure, does it mention robust feature for the first time, they define some math functions for features} \siyuan{I think you may mean "Robustness May Be at Odds with Accuracy" or "Adversarial Examples are not Bugs, they are Features", where robust features are described.}\xz{let's cite both}

Table~\ref{tab:randomized_smoothing} shows the randomized smoothing results on the three attacks. 
We only conduct the experiment on VGG-16 with GTSRB, using 3000 samples and  $\alpha = 0.001$.
We study multiple Gaussian noise variance $\sigma$ settings.
%, meaning confidence $> 99.9\%$. For every trigger with every Gaussian noise variance $\sigma$, 
For each setting, we report the certified radius $R$ and the accuracy for smoothed classifier $g$ on both the benign and the malicious samples. From the results, when $\sigma$ is normal, 
the certified radius $R$ for malicious samples tends to be smaller than that of benign samples but the differences are  within a normal range. The accuracy is almost equally high.
This demonstrates the robustness of these attacks, including DFST, in the context of random smoothing.
%However, the radius differences still fall into a normal variation range, dem
%images with their ground truth labels and trojan images with their target attacking labels.

%From the table, it is hard to tell the difference between benign and trojan data, since they have similar $R$ on each situation, even for the most fragile linear filter triggers.

%both adversarial training on trojan attack model of three trigger types. All the three model is VGG and the experimenting dataset is CIFAR-10. The red rows mean the attack success rate, represented by testing accuracy of trojaned images. It's clear that DFST and watermark attack trigger could sustain both PGD and FGSM adversarial training, while attack accuracy of linear filter drops greatly, which proves that both the watermark attack and our DFST are robust against simple adversarial training.

%\siyuan{Add one more.}
Table~\ref{tab:rethinking} shows the results of  pre-processing defense~\cite{li2020rethinking} on DFST attacked models. We conduct experiments on VGG and ResNet32 with GTSRB using 6 transformation algorithms, Flip (flipping), ShrinkPad (random padding after shrinking with the parameters meaning the number of shrunk pixels), Gaussian (adding Gaussian noise with 0 mean and some std values), Brightness (changing brightness), Saturation (changing saturation), and Contrast (changing contrast). Note that we change the  images' brightness, saturation and contrast using the image enhancing functions in PIL ({\it PIL.ImageEnhance.Brightness(img).enhance(extent)} ).
We report the clean accuracy and attack success rate (ASR) changes on all the transformation algorithms. The results show that DFST is robust as ASR does not degrade in any transformation even when the clean accuracy degrades.

% Please add the following required packages to your document preamble:
% \usepackage[table,xcdraw]{xcolor}
% If you use beamer only pass "xcolor=table" option, i.e. \documentclass[xcolor=table]{beamer}
\begin{table*}[]
\centering
\caption{Attack robustness after adversarial training. ``Benign Test'' presents the accuracy of benign inputs and ``Attack Test" presents the attack success rate on samples with triggers.} \label{tab:robustness}
\begin{tabular}{ccccccccc}
{\color[HTML]{333333} Dataset} & {\color[HTML]{333333} Model} & {\color[HTML]{333333} Trigger}       & \multicolumn{2}{c}{{\color[HTML]{333333} Without Adv Training}} & \multicolumn{2}{c}{{\color[HTML]{333333} PGD L-inf 5/255}}        & \multicolumn{2}{c}{{\color[HTML]{333333} FGSM Stride 0.2}}        \\
                               &                              &                                      & Benign Test                    & Attack Test                    & Benign Test                  & {\color[HTML]{FE0000} Attack Test} & Benign Test                  & {\color[HTML]{FE0000} Attack Test} \\ \hline
{\color[HTML]{333333} }        & {\color[HTML]{333333} }      & {\color[HTML]{333333} Watermark}     & {\color[HTML]{333333} 0.916}   & {\color[HTML]{333333} 0.988}   & {\color[HTML]{333333} 0.869} & {\color[HTML]{FE0000} 0.996}       & {\color[HTML]{333333} 0.913} & {\color[HTML]{FE0000} 0.992}       \\
{\color[HTML]{333333} }        & {\color[HTML]{333333} NiN}   & {\color[HTML]{333333} Linear Filter} & {\color[HTML]{333333} 0.913}   & {\color[HTML]{333333} 0.963}   & {\color[HTML]{333333} 0.864} & {\color[HTML]{FE0000} 0.867}       & {\color[HTML]{333333} 0.911} & {\color[HTML]{FE0000} 0.894}       \\
{\color[HTML]{333333} }        & {\color[HTML]{333333} }      & {\color[HTML]{333333} DFST}          & {\color[HTML]{333333} 0.899}   & {\color[HTML]{333333} 0.970}   & {\color[HTML]{333333} 0.855} & {\color[HTML]{FE0000} 0.963}       & {\color[HTML]{333333} 0.902} & {\color[HTML]{FE0000} 0.973}       \\ \cline{2-9} 
{\color[HTML]{333333} }        & {\color[HTML]{333333} }      & {\color[HTML]{333333} Watermark}     & {\color[HTML]{333333} 0.925}   & {\color[HTML]{333333} 0.999}   & {\color[HTML]{333333} 0.878} & {\color[HTML]{FE0000} 0.999}       & {\color[HTML]{333333} 0.918} & {\color[HTML]{FE0000} 1.000}       \\
{\color[HTML]{333333} CIFAR10} & {\color[HTML]{333333} VGG}   & {\color[HTML]{333333} Linear Filter} & {\color[HTML]{333333} 0.925}   & {\color[HTML]{333333} 0.928}   & {\color[HTML]{333333} 0.879} & {\color[HTML]{FE0000} 0.633}       & {\color[HTML]{333333} 0.918} & {\color[HTML]{FE0000} 0.848}       \\
{\color[HTML]{333333} }        & {\color[HTML]{333333} }      & {\color[HTML]{333333} DFST}          & {\color[HTML]{333333} 0.928}   & {\color[HTML]{333333} 0.999}   & {\color[HTML]{333333} 0.858} & {\color[HTML]{FE0000} 0.990}       & {\color[HTML]{333333} 0.902} & {\color[HTML]{FE0000} 0.985}       \\ \cline{2-9} 
                               &                              & Watermark                            & 0.921                          & 0.994                          & 0.839                        & {\color[HTML]{FE0000} 0.999}       & 0.913                        & {\color[HTML]{FE0000} 0.992}       \\
                               & ResNet                       & Linear Filter                        & 0.921                          & 0.960                          & 0.836                        & {\color[HTML]{FE0000} 0.843}       & 0.911                        & {\color[HTML]{FE0000} 0.910}       \\
                               &                              & DFST                                 & 0.900                          & 0.976                          & 0.827                        & {\color[HTML]{FE0000} 0.943}       & 0.889                        & {\color[HTML]{FE0000} 0.972}       \\ \hline
                               &                              & Watermark                            & 0.984                          & 1.000                          & 0.964                            & {\color[HTML]{FE0000} 1.000}           & 0.980                        & {\color[HTML]{FE0000} 1.000}       \\
                               & NiN                          & Linear Filter                        & 0.969                          & 0.994                          & 0.966                        & {\color[HTML]{FE0000} 0.424}       & 0.975                        & {\color[HTML]{FE0000} 0.954}       \\
                               &                              & DFST                                 & 0.963                          & 0.988                          & 0.964                        & {\color[HTML]{FE0000} 0.981}       & 0.975                        & {\color[HTML]{FE0000} 0.981}       \\ \cline{2-9} 
                               &                              & Watermark                            & 0.983                          & 1.000                          & 0.944                            & {\color[HTML]{FE0000} 1.000}
                               & 0.968                        & {\color[HTML]{FE0000} 1.000}       \\
GTSRB                          & VGG                          & Linear Filter                        & 0.970                          & 0.993                          & 0.946                        & {\color[HTML]{FE0000} 0.439}       & 0.975                        & {\color[HTML]{FE0000} 0.991}       \\
                               &                              & DFST                                 & 0.962                          & 0.987                          & 0.957                        & {\color[HTML]{FE0000} 0.971}       & 0.976                        & {\color[HTML]{FE0000} 0.964}       \\ \cline{2-9} 
                               &                              & Watermark                            & 0.969                          & 1.000                          & 0.966                        & {\color[HTML]{FE0000} 1.000}       & 0.982                        & {\color[HTML]{FE0000} 1.000}       \\
                               & ResNet                       & Linear Filter                        & 0.971                          & 0.992                          & 0.963                        & {\color[HTML]{FE0000} 0.416}       & 0.976                        & {\color[HTML]{FE0000} 0.991}       \\
                               &                              & DFST                                 & 0.967                          & 0.980                          & 0.964                        & {\color[HTML]{FE0000} 0.985}       & 0.975                        & {\color[HTML]{FE0000} 0.998}       \\ \hline
\end{tabular}
\end{table*}

% Please add the following required packages to your document preamble:
% \usepackage{multirow}
\begin{table*}[]
\centering
%\large
\caption{Randomized smoothing results on trojaned VGG16 on GTSRB} \label{tab:randomized_smoothing}
\begin{tabular}{cc|cc|cc|cc|cc}
\multirow{2}*{Dataset}                            &          & \multicolumn{2}{c|}{$\sigma = 0.5$} & \multicolumn{2}{c|}{$\sigma = 0.2$} & \multicolumn{2}{c|}{$\sigma = 0.1$} & \multicolumn{2}{c}{$\sigma = 0.05$}      \\ \cline{3-10} 
~                                   &          & Benign         & Trojan        & Benign         & Trojan        & Benign         & Trojan        & Benign                     & Trojan \\ \hline
\multicolumn{1}{c|}{DFST}          & Radius   & 0.198          & 0.304         & 0.292          & 0.160         & 0.237          & 0.297         & \multicolumn{1}{c|}{0.270} & 0.163  \\ \cline{2-10} 
\multicolumn{1}{c|}{}              & Accuracy & 0.728          & 0.931         & 0.977          & 0.977         & 1.000          & 1.000         & \multicolumn{1}{c|}{1.000} & 1.000  \\ \hline
\multicolumn{1}{c|}{Watermark}     & Radius   & 0.175          & 0.247         & 0.238          & 0.151         & 0.763          & 0.291         & \multicolumn{1}{c|}{0.195} & 0.143  \\ \cline{2-10} 
\multicolumn{1}{c|}{}              & Accuracy & 0.488          & 0.898         & 0.929          & 0.934         & 1.000          & 1.000         & \multicolumn{1}{c|}{1.000} & 1.000  \\ \hline
\multicolumn{1}{c|}{Linear Filter} & Radius   & 0.153          & 0.197         & 0.201          & 0.155         & 0.749          & 0.303         & \multicolumn{1}{c|}{0.163} & 0.098  \\ \cline{2-10} 
\multicolumn{1}{c|}{}              & Accuracy & 0.389          & 0.913         & 0.929          & 0.930         & 0.966          & 0.806         & \multicolumn{1}{c|}{0.789} & 0.801 
\end{tabular}
\end{table*}

\begin{table*}[]
\centering
\caption{Results of~\cite{li2020rethinking} on trojaned VGG and ResNet32 on GTSRB} \label{tab:rethinking}
\begin{tabular}{ccccc}
\multirow{2}*{Transformation}     & \multicolumn{2}{c}{VGG} & \multicolumn{2}{c}{ResNet32} \\ \cline{2-5} 
~                   & Clean       & ASR       & Clean         & ASR          \\ \hline
Standard           & 0.92        & 0.99      & 0.92          & 0.99         \\ \hline
Flip               & 0.92        & 0.98      & 0.92          & 0.99         \\ \hline
ShrinkPad-1        & 0.91        & 0.98      & 0.90          & 0.98         \\
ShrinkPad-2        & 0.90        & 0.97      & 0.90          & 0.96         \\
ShrinkPad-4        & 0.88        & 0.93      & 0.87          & 0.88         \\ \hline
Gaussian-std=5/255 & 0.89        & 0.98      & 0.88          & 0.86         \\
Gaussian-std=5/255 & 0.57        & 0.95      & 0.65          & 0.71         \\ \hline
Brightness=0.8     & 0.92        & 0.98      & 0.91          & 0.98         \\
Brightness=1.2     & 0.92        & 0.97      & 0.90          & 0.98         \\ \hline
Saturation=0.8     & 0.92        & 0.98      & 0.92          & 0.97         \\
Saturation=1.2     & 0.92        & 0.98      & 0.91          & 0.99         \\ \hline
Contrast=0.8       & 0.92        & 0.99      & 0.92          & 0.99         \\
Contrast=1.2       & 0.92        & 0.97      & 0.91          & 0.97         \\ \hline
\end{tabular}
\end{table*}